\pdfoutput=1
\RequirePackage{fix-cm}
\RequirePackage{fixltx2e}
\documentclass[a4paper,english,10pt,journal,compsoc,twoside]{IEEEtran}
\usepackage[T1]{fontenc}
\usepackage[latin9]{inputenc}
\usepackage{color}
\usepackage{babel}
\usepackage{units}
\usepackage{amsmath}
\usepackage{amsthm}
\usepackage{amssymb}
\usepackage{graphicx}

\makeatletter

\pdfpageheight\paperheight
\pdfpagewidth\paperwidth

\providecommand{\tabularnewline}{\\}

\theoremstyle{plain}
\newtheorem{thm}{\protect\theoremname}
\theoremstyle{plain}
\newtheorem{prop}[thm]{\protect\propositionname}

\usepackage{ifpdf}

\ifCLASSOPTIONcompsoc
  \usepackage[nocompress]{cite}
\else
  \usepackage{cite}
\fi

\usepackage{array}
\usepackage{mdwmath}
\usepackage{mdwtab}
\usepackage{eqparbox}

\ifCLASSOPTIONcompsoc
  \usepackage[caption=false,font=footnotesize,labelfont=sf,textfont=sf]{subfig}
\else
  \usepackage[caption=false,font=footnotesize]{subfig}
\fi

\let\oldmarginpar\marginpar
\renewcommand\marginpar[1]{\-\oldmarginpar[\raggedleft\footnotesize #1]%
{\raggedright\footnotesize #1}}

\usepackage{layouts} 

\usepackage{algpseudocode}

\usepackage{afterpage}
\usepackage{flushend}

\renewcommand{\mid}{\;\ifnum\currentgrouptype=16 \middle\fi|\;}

\usepackage{setspace}
\let\oldcases\cases
\let\oldendcases\endcases
\renewenvironment{cases}{\setstretch{0.75}\oldcases}{\oldendcases}

\@ifundefined{showcaptionsetup}{}{%
 \PassOptionsToPackage{caption=false}{subfig}}
\usepackage{subfig}
\makeatother

\providecommand{\propositionname}{Prop.}
\providecommand{\theoremname}{Theorem}

\usepackage[unicode=true,
 bookmarks=true,bookmarksnumbered=true,bookmarksopen=false,
 breaklinks=true,pdfborder={0 0 0},pdfborderstyle={},backref=false,colorlinks=true]
 {hyperref}
\hypersetup{pdftitle={Kernel Density Estimation-Based Markov Models with Hidden State},
 pdfauthor={Gustav Eje Henter, Arne Leijon, and W. Bastiaan Kleijn},
 pdfkeywords={hidden Markov models, nonparametric methods, kernel density estimation, autoregressive models, time-series bootstrap},
 citecolor=blue}
\usepackage{breakurl}

\begin{document}
\ifCLASSOPTIONpeerreview
\else
  \markboth{Preprint. Work in progress.}{Preprint. Work in progress.}%
\fi

\IEEEtitleabstractindextext{%
\begin{abstract}
We consider Markov models of stochastic processes where the next-step
conditional distribution is defined by a kernel density estimator
(KDE), similar to Markov forecast densities and certain time-series
bootstrap schemes. The KDE Markov models (KDE-MMs) we discuss are
nonlinear, nonparametric, fully probabilistic representations of stationary
processes, based on techniques with strong asymptotic consistency
properties. The models generate new data by concatenating points from
the training data sequences in a context-sensitive manner, together
with some additive driving noise. We present novel EM-type maximum-likelihood
algorithms for data-driven bandwidth selection in KDE-MMs. Additionally,
we augment the KDE-MMs with a hidden state, yielding a new model class,
KDE-HMMs. The added state variable captures non-Markovian long memory
and signal structure (e.g., slow oscillations), complementing the
short-range dependences described by the Markov process. The resulting
joint Markov and hidden-Markov structure is appealing for modelling
complex real-world processes such as speech signals. We present guaranteed-ascent
EM-update equations for model parameters in the case of Gaussian kernels,
as well as relaxed update formulas that greatly accelerate training
in practice. Experiments demonstrate increased held-out set probability
for KDE-HMMs on several challenging natural and synthetic data series,
compared to traditional techniques such as autoregressive models,
HMMs, and their combinations.
\end{abstract}

}

\title{Kernel Density Estimation-Based\\
Markov Models with Hidden State}

\author{Gustav~Eje Henter,~\IEEEmembership{Member,~IEEE}, Arne Leijon,~\IEEEmembership{Member,~IEEE},\\
and W.~Bastiaan Kleijn,~\IEEEmembership{Fellow,~IEEE}%
\IEEEcompsocitemizethanks{\IEEEcompsocthanksitem G.\ E.\ Henter is with the Department of Speech, Music and Hearing, KTH Royal Institute of Technology, Stockholm, Sweden. This research took place while he was with the School of Electrical Engineering at the same university. E-mail: \protect\href{mailto:ghe@kth.se}{\nolinkurl{ghe@kth.se}}.%
\IEEEcompsocthanksitem A.\ Leijon recently retired from the School of Electrical Engineering, KTH Royal Institute of Technology, Stockholm, Sweden. E-mail: \protect\href{mailto:leijon@kth.se}{\nolinkurl{leijon@kth.se}}.%
\IEEEcompsocthanksitem W.\ B.\ Kleijn is with the School of Engineering and Computer Science, Victoria University of Wellington, New Zealand, and the Circuits and Systems Group, Delft University of Technology, The Netherlands. This research took place while he was with the School of Electrical Engineering, KTH Royal Institute of Technology, Stockholm, Sweden.\protect\\%
E-mail: \protect\href{mailto:bastiaan.kleijn@ecs.vuw.ac.nz}{\nolinkurl{bastiaan.kleijn@ecs.vuw.ac.nz}}.%
\IEEEcompsocthanksitem This research was supported by the LISTA (Listening Talker) project. The project LISTA acknowledges the financial support of the Future and Emerging Technologies (FET) programme within the Seventh Framework Programme for Research of the European Commission, under FET-Open grant number: 256230.}%
\thanks{Manuscript last revised November 16, 2016.\protect\\%
(Corresponding author: Gustav Eje Henter.)}}

\maketitle
\IEEEdisplaynotcompsoctitleabstractindextext
\IEEEpeerreviewmaketitle

\ifCLASSOPTIONcompsoc
  \IEEEraisesectionheading{\section{Introduction}}\label{paperd:sec:introduction}\par
\else
  \section{Introduction}\label{paperd:sec:introduction}\par
\fi
\IEEEPARstart{T}{ime} series and other sequence data are ubiquitous in nature. To recognize
patterns or make decisions based on observations in the face of uncertainty
and natural variation, or to generate new data, e.g., for synthesizing
speech, we need capable stochastic models.

Exactly what model to use depends on the situation. The standard approach
is to propose a mathematical framework, and then let data fill in
the unknowns by estimating parameters \cite{bishop2006a,henter2013probabilistic}.
If the data-generating process is well understood, it may be possible
to write down an appropriate model form directly. If not, we must
fall back on the extensive library of general-purpose statistical
models available. With little data, simple, tried-and-true approaches
are typically used. These tend to involve many assumptions on the
nature of the data-generating process. Model accuracy may suffer if
these assumptions are incorrect. Once more data becomes available,
such as in today's ``big data'' paradigm, it is often possible to
get a better description by fitting more advanced models with additional
parameters. Using a complex model is certainly no guarantee for better
results, however, and finding out which particular description to
use tends to be a laborious trial-and-error process.

In this article, we consider a class of nonparametric models for discrete-time,
continuous-valued stationary data series, where conditional next-step
distributions are defined by kernel density estimators. Unlike standard
parametric techniques, these models can converge on a significantly
broader class of ergodic finite-order Markov processes as the training
data material grows large. This makes the models widely applicable,
and is especially compelling for data where the generating process
is complex and nonlinear, or otherwise poorly understood.

Our main contributions are 1) extending the nonparametric Markov models
with a discrete-valued hidden state, and 2) presenting several guaranteed-ascent
iterative update formulas for maximum-likelihood parameter estimation,\footnote{\emph{Nonparametric models} can be defined as sets of probability
distributions indexed by a parameter $\boldsymbol{\theta}\in\boldsymbol{\Theta}$,
where the dimensionality of the space $\boldsymbol{\Theta}$ grows
with the amount of data; hence a nonparametric model still has parameters
that may need to be estimated.} applicable both with and without hidden state. The added state variable
allows models to capture long-range dependences, patterns with variable
duration, and similar structure, on top of the short-range correlations
described by the Markov model. This is shown to improve finite-sample
performance on challenging real-world data. The state variable can
also be used as an input signal to control process output, and is
attractive for recognition and synthesis applications, particularly
with speech signals, where the hidden states can be identified with
language units such as phones.

The remainder of the paper is organized as follows: Section \ref{paperd:sec:background}
discusses established techniques for modelling Markovian processes
with and without hidden state, along with sample applications. Section
\ref{paperd:sec:models} then introduces kernel density-based time
series models and their properties. Parameter estimation is subsequently
treated in Section \ref{paperd:sec:training}. Sections \ref{paperd:sec:synthexperiments}
and \ref{paperd:sec:realexperiments} present experimental results,
while Section \ref{paperd:sec:conclusion} concludes.

\section{Background\label{paperd:sec:background}}

In this section we discuss Markov models and hidden-state models for
strictly stationary and ergodic sequence data, and also address how
the two approaches can be combined to describe more complex natural
processes.

\subsection{Markov Models}

Let the underline notation $\underline{X}_{1}^{T}=\left(X_{1},\,\ldots,\,X_{T}\right)$
represent a sequence of variables from a stochastic process. A process
satisfying 
\begin{align}
f_{X_{t}\mid\underline{X}_{-\infty}^{t-1}}\left(x_{t}\mid\underline{x}_{-\infty}^{t-1}\right) & \equiv f_{X_{t}\mid\underline{X}_{t-p}^{t-1}}\left(x_{t}\mid\underline{x}_{t-p}^{t-1}\right)\textrm{;}\label{paperd:eq:markovdef}
\end{align}
 is said to be a \emph{Markov process of order $p$}.\footnote{We will use capital letters to denote random variables (RVs), letting
lower-case letters signify observations, i.e., specific, nonrandom
outcomes of RVs.} The relation implies that future evolution is conditionally independent
of the past, given the latest observations\textemdash the \emph{context}
or \emph{state} $\underline{x}_{t-p}^{t-1}$. In other words, the
process has a short, finite memory, and knowing the most recent samples
suffices for optimal prediction. $p=0$ means that variables are independent.
The Bayesian network in Fig.\ \ref{paperd:fig:mm2deps} illustrates
the between-variable dependences when $p=2$.

Continuous-valued Markovian data can be modelled using linear as well
as nonlinear models. Standard linear autoregressive (AR) and autoregressive
moving-average (ARMA) models are perhaps the most well-known. Among
nonlinear models one finds piecewise-linear, regime-switching approaches
such as self-exciting threshold autoregressive\emph{ }(SETAR) models
\cite{tong1980a}, non-recurrent time-delay neural networks (TDNN)
\cite{waibel1989a}, or kernel-based AR models \cite{kallas2011a,kallas2012a}.
Similar to regular AR models, the latter can be made probabilistic
by exciting them with a random process, typically white noise.

All the above models are parametric, and can therefore only converge
on processes in their finite-dimensional parametric family. This may
limit the achievable accuracy. As it turns out, convergent, general-purpose
models are possible using nonparametric techniques. In Section \ref{paperd:sec:models}
we highlight and extend a simple but flexible approach based on kernel
density estimation, drawing on \cite{roussas1969a} and \cite{rajarshi1990a}.
The technique is asymptotically consistent for stationary and ergodic
Markov processes\footnote{Not all nonparametric Markov models have this general convergence
property, the residual bootstrap of \cite{clements1997a} being one
counterexample.} \cite{hyndman1996a} despite having only a single free parameter,
and outperforms traditional methods on multiple datasets.

\subsection{Hidden-State Models\label{paperd:sub:hmms}}

\begin{figure}[tp]
\begin{centering}
\subfloat[\label{paperd:fig:mm2deps}Second-order Markov model.]{\begin{centering}
\includegraphics[width=0.48\columnwidth]{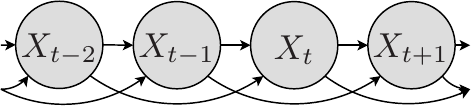}
\par\end{centering}
}
\par\end{centering}
\begin{centering}
\subfloat[\label{paperd:fig:hmmdeps}Model with hidden state.]{\begin{centering}
\includegraphics[width=0.48\columnwidth]{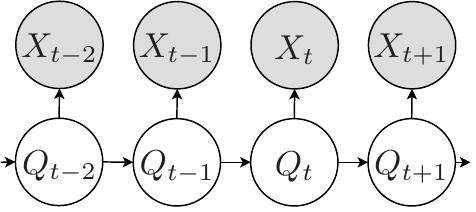}
\par\end{centering}
}\subfloat[\label{paperd:fig:ar2-hmmdeps}Combined model.]{\begin{centering}
\includegraphics[width=0.48\columnwidth]{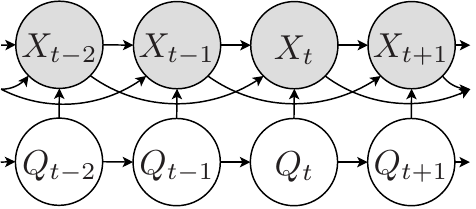}
\par\end{centering}
}
\par\end{centering}
\caption{Variable dependences in various sequence models. $X$s denote observed
process values while $Q$s are unobserved (hidden) state variables.}
\vspace{-2ex}
\end{figure}

To capture long-range dependences, a Markov model may require a high
order, many parameters, and a large dataset to provide a reasonable
description. A more efficient approach is to use models with a hidden,
unobservable state variable $Q_{t}$ that governs time-series evolution.
The state is assumed to follow a Markov process, while the observed
process values, in turn, are stochastic functions $f_{X_{t}\mid Q_{t}}$
of the current state alone. This dependence structure is illustrated
in Fig.\ \ref{paperd:fig:hmmdeps}. In this framework, discrete state
spaces lead to hidden Markov models (HMMs) \cite{rabiner1989a} and
variants thereof, while continuous state spaces yield Kalman filters
\cite{welch2001a} and nonlinear extensions such as those presented
in \cite{wan1997a} and \cite{wan2000a}.

Even though the hidden state $\underline{Q}$ satisfies the Markov
property, the distribution of the observed data $\underline{X}$ typically
cannot be represented by any finite-order Markov process. This is
clear, e.g., from the forward algorithm 
\begin{multline}
f_{Q_{t}\mid\underline{X}_{t_{0}}^{t}}\left(q\mid\underline{x}_{t_{0}}^{t}\right)\propto f_{X_{t}\mid Q_{t}}\left(x_{t}\mid q\right)\\
\cdot\sum_{q'}P\left(Q_{t}=q\mid Q_{t-1}=q'\right)f_{Q_{t-1}\mid\underline{X}_{t_{0}}^{t-1}}\left(q'\mid\underline{x}_{t_{0}}^{t-1}\right)\label{paperd:eq:forwardalgo}
\end{multline}
 for state inference in HMMs \cite{rabiner1989a}, which recursively
depends on all previous samples. Hidden-state models can thus integrate
predictive information over arbitrary lengths of time, providing efficient
representations of long-range signal structure such as, for instance,
the order of sounds in speech signals or different chart patterns
in financial technical analysis. At the same time, efficient inference
is possible for HMMs using the forward-backward algorithm \cite{rabiner1989a}.

Like the components in a mixture model, the discrete state of hidden-Markov
models essentially partitions the data into different subsets, each
explained by a different distribution. This can capture more detail
in the data distribution than can a single component. HMMs furthermore
model how components correlate across time, making them capable of
representing series data.

\subsection{Combined Models\label{paperd:sub:combinationmodels}}

Real-world data series often exhibit both short-range correlations
and long-range structure. In theory, this can be represented by a
discrete-state HMM given a sufficient number of states, but in practice
that number may be prohibitively large. We see that HMMs trade the
limited memory \emph{range} of Markov models for a limited memory
\emph{resolution} (a finite state space).

To get the best of both worlds, Markovian observation dependences
and hidden-state dynamics can be combined within a single model. This
results in a model structure as illustrated in Fig.\ \ref{paperd:fig:ar2-hmmdeps},
where between-sample dependences follow a Markov process with properties
determined by the hidden state. Such a structure is seen in, e.g.,
GARCH models \cite{bollerslev1986a} from econometrics, or in SETAR
models driven by an unobservable Markov chain (so-called Markov switching
models \cite{hamilton1989a,tong1980a}). The addition of \emph{dynamic
features} (velocity and acceleration) to ordinary HMMs\textemdash a
common practice in acoustic modelling of speech \cite{dines2009a}\textemdash can
be viewed as a method for adding implicit between-frame correlations.
Autoregressive HMMs (AR-HMMs) \cite{shannon2009a} and trajectory
HMMs \cite{zen2007b} are similar models that make these dependences
explicit. We will use AR-HMMs as a baseline since they are based on
standard AR models and have a directed structure that allows efficient
sampling and parameter estimation.

The Markovian part in a combined model typically captures simple and
predictable local aspects of the data, such as continuity, allowing
the hidden state to concentrate on more complex, long-range interactions.
In acoustic modelling of speech, e.g., \cite{zen2007b,zen2009a},
dynamic features capture correlations between analysis frames due
to physical constraints on the motion of speech articulators, while
the hidden states are based on a language model (e.g., $n$-grams)
and account for grammar and other large-scale language structures.

As a major contribution of this article, we extend kernel density-based
Markov models as used in \cite{rosenblatt1969a,rajarshi1990a} with
a hidden state. This is found to improve the distribution-prediction
performance for held-out data over comparable parametric approaches
such as AR-HMMs.

\section{KDE Models for Time Series\label{paperd:sec:models}}

In this section, we first review kernel density estimation of probability
distributions, and how it can be adapted to describe and generate
Markovian processes. Novel models with hidden state are introduced
in Section \ref{paperd:sub:kde-hmm}.

From now on, we will use the letters $y$ and $x$ to denote training
data and test data, respectively. $\mathcal{D}$ represents a set
of training data, e.g., a training data sequence $\mathcal{D}=\underline{y}_{1}^{N}$,
while boldface signifies vector-valued variables.

\subsection{Kernel Density Estimation}

\emph{Kernel density estimation} (KDE) or \emph{Parzen estimation},
introduced in \cite{rosenblatt1956a,parzen1962a} and discussed in
more detail in \cite{silverman1986a}, is a nonparametric method for
estimating a $D$-dimensional probability density $f_{\boldsymbol{X}}$
from a finite sample $\mathcal{D}=\left\{ \boldsymbol{y}_{n}\right\} _{n=1}^{N}$,
$\boldsymbol{y}_{n}\in\mathbb{R}^{D}$, by convolving the empirical
density function with a kernel function $k\left(\boldsymbol{r}\right)$.
The resulting estimated pdf can be written as 
\begin{align}
\widehat{f}_{\boldsymbol{X}}\left(\boldsymbol{x}\mid\mathcal{D};\,h\right) & =\frac{1}{N}\sum_{n=1}^{N}\frac{1}{h^{D}}k\left(\frac{1}{h}\left(\boldsymbol{x}-\boldsymbol{y}_{n}\right)\right)\textrm{,}\label{paperd:eq:kdehpdf}
\end{align}
 where the \emph{bandwidth} $h>0$ is a scale parameter adjusting
the width of the kernel.

We see that the KDE in (\ref{paperd:eq:kdehpdf}) always defines a
proper density if we require that $k\left(\boldsymbol{r}\right)\geq0$
and that $k\left(\boldsymbol{r}\right)$ integrates to one. For technical
reasons we also require the first moment $\int\boldsymbol{r}k\left(\boldsymbol{r}\right)\mathrm{d}\boldsymbol{r}$
to be zero and that the second moment $\int\boldsymbol{r}\boldsymbol{r}^{\intercal}k\left(\boldsymbol{r}\right)\mathrm{d}\boldsymbol{r}$
is bounded. A wide variety of kernels exists, with the \emph{squared-exponential}
or \emph{Gaussian kernel} 
\begin{align}
k\left(\boldsymbol{r}\right) & =\frac{1}{\sqrt{2\pi}}\exp\left(-\frac{1}{2}\boldsymbol{r}^{\intercal}\boldsymbol{r}\right)\label{paperd:eq:gausskern}
\end{align}
 being a common example. For simplicity, we generally assume that
the kernel factors across dimensions as 
\begin{align}
k\left(\boldsymbol{r}\right) & =\prod_{d=1}^{D}k_{d}\left(r_{d}\right)\textrm{;}
\end{align}
 this is known as a \emph{product kernel}. Furthermore, we often take
all component functions $k_{d}\left(\cdot\right)$ to be identical.

The bandwidth $h$ controls the degree of smoothing the KDE applies
to the empirical distribution function. \emph{Bandwidth selection},
the task of choosing this $h$, is key to KDE performance. With an
appropriate adaptive bandwidth selection scheme, such that $h\to0$
while $Nh^{D}\to\infty$ as $N\to\infty$, KDE can be shown to converge
asymptotically on the true probability density function in a mean-squared
error sense, regardless of $k$. (Kernel shape $k\left(\cdot\right)$
is less important, with many standard choices yielding near-optimal
performance \cite{wand1995a}.) Essentially, given an ever-growing
collection of progressively more localized basis functions with centres
drawn from $f_{\boldsymbol{X}}$, we can eventually represent arbitrarily
small details in the pdf. Moreover, KDE attains the best possible
convergence rate for nonparametric estimators \cite{wahba1975a},
assuming optimal bandwidth selection.

In practice, KDE finite-sample performance depends heavily on the
data and the particular bandwidth chosen. Smooth distributions are
typically easy to learn and can use large $h$, whereas complicated
distributions require more data and narrow bandwidths to bring out
the details. Consequently, there are many techniques for choosing
the bandwidth in a data-driven manner: see, e.g., \cite{cau1994a}
or \cite{turlach1993a} for reviews.

\subsection{Kernel Conditional Density Estimation}

We now consider how to approximate Markov processes. Using the Markov
property (\ref{paperd:eq:markovdef}), the density function for a
univariate sequence $\underline{x}_{1}^{T}$ from a general stationary
nonlinear Markov process of order $p$ can be written 
\begin{equation}
f_{\underline{X}_{1}^{T}}(\underline{x}_{1}^{T})=f_{\underline{X}_{1}^{p}}\left(\underline{x}_{1}^{p}\right)\prod_{t=p+1}^{T}f_{X_{p+1}\mid\underline{X}_{1}^{p}}\left(x_{t}\mid\underline{x}_{t-p}^{t-1}\right)\textrm{.}\label{paperd:eq:generalmarkov}
\end{equation}
 It is sufficient to specify the stationary conditional next-step
distribution $f_{X_{p+1}\mid\underline{X}_{1}^{p}}$ to uniquely determine
the $x$-process and its associated stationary distribution $f_{\underline{X}_{1}^{p}}$.
A key idea in this paper is to use KDEs, specifically \emph{kernel
conditional density estimation} (KCDE), to estimate this conditional
distribution from data.

Given two variables $\boldsymbol{X}\in\mathbb{R}^{D}$ and $\boldsymbol{X}'\in\mathbb{R}^{D'}$
and a kernel density estimate $\widehat{f}_{\boldsymbol{X},\,\boldsymbol{X}'}\left(\boldsymbol{x},\,\boldsymbol{x}'\right)$
of their joint distribution $f_{\boldsymbol{X},\,\boldsymbol{X}'}$
from training data pairs $\left\{ \left(\boldsymbol{y}_{n},\,\boldsymbol{y}'_{n}\right)\right\} _{n=1}^{N}$,
the estimate $\widehat{f}_{\boldsymbol{X},\,\boldsymbol{X}'}$ also
induces a conditional distribution 
\begin{align}
\widehat{f}_{\boldsymbol{X}\mid\boldsymbol{X}'}\left(\boldsymbol{x}\mid\boldsymbol{x}';\,h,\,h'\right)=\frac{\widehat{f}_{\boldsymbol{X},\,\boldsymbol{X}'}\left(\boldsymbol{x},\,\boldsymbol{x}';\,h,\,h'\right)}{\int\widehat{f}_{\boldsymbol{X},\,\boldsymbol{X}'}\left(\boldsymbol{\xi},\,\boldsymbol{x}';\,h,\,h'\right)\mathrm{d}\boldsymbol{\xi}}\\
=\sum_{n=1}^{N}\frac{k_{\boldsymbol{X}'}\left(\frac{\boldsymbol{x}^{\prime}-\boldsymbol{y}_{n}^{\prime}}{h'}\right)}{\sum_{n'=1}^{N}k_{\boldsymbol{X}'}\left(\frac{\boldsymbol{x}^{\prime}-\boldsymbol{y}_{n'}^{\prime}}{h'}\right)}\frac{1}{h^{D}}k_{\boldsymbol{X}}\left(\frac{\boldsymbol{x}-\boldsymbol{y}_{n}}{h}\right) & \textrm{;}\label{paperd:eq:kcdedef}
\end{align}
 this is the KCDE for $f_{\boldsymbol{X}\mid\boldsymbol{X}'}$ \cite{rosenblatt1969a,hyndman1996a,fan1996a}.
(We have assumed that the kernel factors between $\boldsymbol{X}$
and $\boldsymbol{X}'$.)

The estimator in (\ref{paperd:eq:kcdedef}) is (MSE) consistent if
bandwidths satisfy $h,\,h'\to0$ and $Nh^{D}h^{\prime D'}\to\infty$
when $N\to\infty$. As always, this flexibility and consistency comes
at a price of increased computational complexity and memory demands
over parametric approaches, since all $N$ data points typically must
be stored and used in calculations, making, e.g., bandwidth selection
scale as $N^{2}$. Fortunately, approximate $k$-nearest neighbour
techniques can be used to accelerate KDE computations. Reference \cite{holmes2007a}
describes a method based on dual trees yielding speedup factors up
to six orders of magnitude for KDE likelihood computation on 10,000
points from a nine-dimensional census dataset.

\subsection{KDE Markov Models}

Let $\underline{y}_{1}^{N}$ be a sampled data sequence (time series)
from a stationary, ergodic Markov process of interest. Since $\underline{y}_{1}^{N}$
can be seen as a set of samples $\mathcal{D}=\left\{ \underline{y}_{n-p}^{n}\right\} _{n=p+1}^{N}$
from the stationary distribution of $\underline{X}_{1}^{p+1}$, we
can apply KCDE to estimate the conditional next-step distribution
in (\ref{paperd:eq:generalmarkov}). Restricting ourselves to product
kernels where all $k_{d}\left(\cdot\right)$ are identical and use
the same bandwidth (this is appealing since $\widehat{f}_{\underline{X}_{1}^{p+1}}$
is stationary), this KCDE next-step distribution takes the form 
\begin{align}
\widehat{f}_{X_{p+1}\mid\underline{X}_{1}^{p}}\left(x_{t}\mid\underline{x}_{t-p}^{t-1};\,h\right) & =\frac{1}{h}\frac{\sum_{n=p+1}^{N}\prod_{l=0}^{p}k\left(\frac{x_{t-l}-y_{n-l}}{h}\right)}{\sum_{n=p+1}^{N}\prod_{l=1}^{p}k\left(\frac{x_{t-l}-y_{n-l}}{h}\right)}\textrm{.}\label{paperd:eq:kde-mm}
\end{align}
 (Sequences of vectors are a straightforward extension.)

Eq.\ (\ref{paperd:eq:kde-mm}) defines a Markov model which approximates
the data-generating process. We will call this construction a \emph{KDE
Markov model} (KDE-MM), and will write $\widehat{X}$ to distinguish
variables generated from the KDE-MM next-step distribution $\widehat{f}_{X_{p+1}\mid\underline{X}_{1}^{p}}$
in (\ref{paperd:eq:kde-mm}) from $X$, data distributed according
to the reference process defined by $f_{X_{p+1}\mid\underline{X}_{1}^{p}}$
in (\ref{paperd:eq:generalmarkov}).\footnote{\label{paperd:fn:kde-mmperiodic}Although by definition $f_{\widehat{X}_{p+1}\mid\underline{\widehat{X}}_{1}^{p}}\equiv\widehat{f}_{X_{p+1}\mid\underline{X}_{1}^{p}}$,
the stationary distribution $f_{\underline{\widehat{X}}_{1}^{p+1}}$
of $\widehat{X}$ induced by (\ref{paperd:eq:kde-mm}) need not necessarily
match the original KDE pdf $\widehat{f}_{\underline{X}_{1}^{p+1}}$.
This is because $\widehat{f}_{\underline{X}_{1}^{p+1}}$ typically
cannot be a stationary distribution, as its marginal distributions
$\widehat{f}_{X_{l}}$ for $l\in\left\{ 1,\,\ldots,\,p+1\right\} $
are based on different sets of kernel centres (datapoints) $\left\{ y_{n+l}\right\} _{n=1}^{N-p}$
and thus are unlikely to be identical.

If we wish to ensure that $f_{\underline{\widehat{X}}_{1}^{p+1}}\equiv\widehat{f}_{\underline{X}_{1}^{p+1}}$,
it is sufficient to perform periodic extension of the data series
for non-positive indices, so that $y_{-n}\equiv y_{N-n}$ (reminiscent
of periodic extension in Fourier analysis), and change the summations
over $n$ in (\ref{paperd:eq:kde-mm}) to start at $n=1$ rather than
at $p+1$. This makes all marginal distributions $\widehat{f}_{X_{l}}$
identical, though it may introduce out-of-character behaviour into
$\widehat{X}$ in case the beginning and end of the training sequence
do not match up well. Stationarity is easily verified by computing
$\widehat{f}_{\underline{X}_{2}^{p+1}}=\int\widehat{f}_{X_{p+1}\mid\underline{X}_{1}^{p}}\widehat{f}_{\underline{X}_{1}^{p}}\mathrm{d}x_{1}$.

We have implemented the periodic extension for all KDE-MMs (but not
KDE-HMMs) in our experiments. However, (\ref{paperd:eq:kde-mm}) will
always define a proper stationary stochastic process even if this
is not done.}

The use of conditional distributions based on KDE to describe first-order
($p=1$) Markovian data dates back to \cite{rajarshi1990a}, predating
the KCDE paper \cite{hyndman1996a}. At $p=1$, our kernel density-based
Markov model coincides with the next-step distribution in \cite{rajarshi1990a},
assuming the latter uses a product kernel. KDE-MMs can also be seen
as a slight restriction of the \emph{Markov forecast density} (MFD),
introduced for economics by \cite{manzana2008a}, when applied to
one-step prediction (the methods differ in their predictions for multiple
time-steps). The maximum-likelihood-type parameter estimation algorithms
we later present in Section \ref{paperd:sec:training} are new, however.

Using KCDE to describe the next-step conditional distribution has
several advantages over parametric approaches. To begin with, we inherit
the asymptotic consistency properties of KDEs, and the probabilities
$\widehat{f}_{\underline{X}_{1}^{T}}$ of finite substrings in $\underline{\widehat{X}}$
converge on the true probabilities $f_{\underline{X}_{1}^{T}}$ as
$N\to\infty$ (under certain regularity conditions and with appropriately
chosen bandwidths). Moreover, the approach only has a single free
parameter, $h$.

As with other nonparametric models, a potential downside of KDE-MMs
is that per-point computational costs scale linearly with database
size $N$. At the same time, $N$ may need to be large for KDE-based
methods to overtake fast-to-converge but asymptotically biased parametric
models.

\subsection{KDE-MM Data Generation\label{paperd:sub:kde-mmsampling}}

\begin{figure}[tp]
\begin{centering}
\includegraphics[width=0.48\columnwidth]{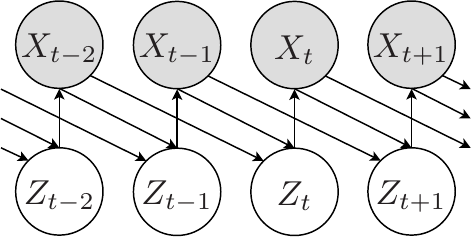}
\par\end{centering}
\caption{\label{paperd:fig:kde2-mmdeps}Variable dependences in a second-order
KDE-MM. The generating component $Z$ is a latent variable.}
\vspace{-2ex}
\end{figure}
Eq.\ (\ref{paperd:eq:kde-mm}) can be rewritten as 
\begin{multline}
\widehat{f}_{X_{p+1}\mid\underline{X}_{1}^{p}}\left(x_{t}\mid\underline{x}_{t-p}^{t-1};\,h\right)\\
=\sum_{n=p+1}^{N}\kappa_{n}\left(\underline{x}_{t-p}^{t-1};\,h\right)\frac{1}{h}k\left(\frac{x_{t}-y_{n}}{h}\right)
\end{multline}
\vspace{-12pt}
\begin{gather}
\kappa_{n}\left(\underline{x}_{t-p}^{t-1};\,h\right)=\frac{\prod_{l=1}^{p}k\left(\frac{x_{t-l}-y_{n-l}}{h}\right)}{\sum_{n'=p+1}^{N}\prod_{l=1}^{p}k\left(\frac{x_{t-l}-y_{n'-l}}{h}\right)}\textrm{,}
\end{gather}
 which can be interpreted as a mixture model with context-dependent
weights. To generate a new datapoint given the context $\underline{x}_{t-p}^{t-1}$,
one selects a mixture component $Z_{t}\in\left\{ p+1,\,\ldots,\,N\right\} $
according to the weights $\kappa_{n}\left(\underline{x}_{t-p}^{t-1};\,h\right)$.
For KDE-MMs this component $z_{t}$ corresponds to an index into the
database $\mathcal{D}$, identifying an exemplar $y_{z_{t}}$ upon
which we base the sample at $t$. $x_{t}$ is then generated from
$y_{z_{t}}$ with $k$-shaped noise added: 
\begin{align}
P\left(Z_{t}=n\mid\underline{X}_{t-p}^{t-1}=\underline{x}_{t-p}^{t-1};\,h\right) & =\kappa_{n}\left(\underline{x}_{t-p}^{t-1};\,h\right)\\
\widehat{f}_{X_{p+1}\mid Z_{p+1},\,\underline{X}_{1}^{p}}\left(x_{t}\mid n,\,\underline{x}_{t-p}^{t-1};\,h\right) & =\\
\widehat{f}_{X_{p+1}\mid Z_{p+1}}\left(x_{t}\mid n;\,h\right) & =\frac{1}{h}k\left(\frac{x_{t}-y_{n}}{h}\right)\textrm{.}\label{paperd:eq:addknoise}
\end{align}
 This staggered structure is illustrated in Fig.\ \ref{paperd:fig:kde2-mmdeps}
for a second-order model ($p=2$). The procedure is also simple to
implement in software. It is easy to see that 
\begin{align}
\widehat{f}_{X_{p+1}\mid\underline{X}_{1}^{p}}\left(x_{p+1}\mid\underline{x}_{1}^{p}\right) & \leq\max_{y_{n}\in\mathcal{D}}\frac{1}{h}k\left(\frac{x_{p+1}-y_{n}}{h}\right)\textrm{,}
\end{align}
 regardless of context $\underline{x}_{1}^{p}$. This bounds the tails
of $\widehat{f}_{X_{t}}$, so the $\widehat{X}$-process has as many
finite moments as the kernel function does, ensuring stability.

The bandwidth $h$ affects the character of the generated data in
two ways. With a sufficiently narrow bandwidth, and assuming $k$
has exponentially decreasing tails, the weight distribution $\kappa_{n}\left(\underline{x}_{t-p}^{t-1}\right)$
for $z_{t}$ will strongly favour the one $n$ that minimizes $\Vert\underline{x}_{t-p}^{t-1}-\underline{y}_{n-p}^{n-1}\Vert$;
call this component $n^{\star}$. Since the added $k$-shaped noise
in (\ref{paperd:eq:addknoise}) has small variance, we have $x_{t}\approx y_{n^{\star}}$
in the typical case. This makes sampled sequences closely follow long
segments of consecutive values from the training data series. If $h$
is increased, the weight distribution becomes more uniform, and sampled
trajectories will increasingly often switch between different training-data
segments, but the amount of added noise also grows. In Markov forecast
densities and in KDE-HMMs as presented in the next section, the two
roles of context sensitivity and additive noise level are assigned
to separate parameters.

The idea of creating new samples by concatenating old data is reminiscent
of the common block bootstrap for time series \cite{hardle2003a},
albeit more refined, since the KDE-MM preferentially selects pieces
that fit well together; this has advantages for the asymptotic convergence
rate of bootstrap statistics \cite{horowitz2003a}. Indeed, both \cite{rajarshi1990a}
and \cite{manzana2008a} apply a bootstrap perspective when introducing
KDE-based methods. The KDE-MM approach is also similar to waveform-similarity-based
overlap-add techniques (WSOLA) \cite{verhelst1993overlap} from signal
processing, although these do not model probability distributions.

\subsection{KDE Hidden Markov Models\label{paperd:sub:kde-hmm}}

Being Markov models, KDE-MMs capture short-range correlations in data
series, but are not well-equipped to describe long-range memory or
structure such as the sequential order of sounds in a speech utterance,
as discussed in Section \ref{paperd:sec:background}. To remedy this,
we introduce a hidden (unobservable), discrete state variable $Q_{t}\in\left\{ 1,\,\ldots,\,M\right\} $
to the model in (\ref{paperd:eq:kde-mm}). $Q_{t}$ is governed by
a first-order Markov chain. This results in hidden Markov models where
the state-conditional output distributions are given by KDEs, specifically
KDE-MMs. In other words, the current state $Q_{t}$ determines which
of a set of $M$ KDE-MMs is used to generate the next sample value
of the process. This construction is completely analogous to the setup
of AR-HMMs \cite{shannon2009a}, but with KDE-MMs instead of linear
AR models. We call the new models \emph{kernel density hidden Markov
models}, or \emph{KDE-HMMs}.

In this paper, we will work with models that, given the current state
$Q_{t}=q$ and context $\underline{x}_{t-p}^{t-1}$, have the form
\begin{multline}
\widehat{f}_{X_{t}\mid Q_{t},\,\underline{X}_{t-p}^{t-1}}\left(x_{t}\mid q,\,\underline{x}_{t-p}^{t-1};\,\boldsymbol{K}\right)\\
=\sum_{n=p+1}^{N}\kappa_{qn}\left(\underline{x}_{t-p}^{t-1};\,\boldsymbol{K}\right)\frac{1}{h_{q0}}k_{q0}\left(\frac{x_{t}-y_{n}}{h_{q0}}\right)\label{paperd:eq:kde-hmm}
\end{multline}
\vspace{-12pt}
\begin{align}
\kappa_{qn}\left(\underline{x}_{t-p}^{t-1};\,\boldsymbol{K}\right) & =\frac{w_{qn}\prod_{l=1}^{p}k_{ql}\left(\frac{x_{t-l}-y_{n-l}}{h_{ql}}\right)}{\sum_{n'=p+1}^{N}w_{qn'}\prod_{l=1}^{p}k_{ql}\left(\frac{x_{t-l}-y_{n'-l}}{h_{ql}}\right)}\textrm{.}\label{paperd:eq:kde-hmmweights}
\end{align}
 $\boldsymbol{K}$ here denotes the set of KDE parameters, $\boldsymbol{K}=\left\{ h_{ql},\,w_{qn}\right\} $
for $q\in\left\{ 1,\,\ldots,\,M\right\} $, $l\in\left\{ 0,\,\ldots,\,p\right\} $,
and $n\in\left\{ p+1,\,\ldots,\,N\right\} $. The function of the
weights $w_{qn}$ is to allow the training data points to be assigned
to different states. This assignment can be either hard (binary) or
soft, but we require $w_{qn}\geq0$ and $\sum_{n=p+1}^{N}w_{qn}=1$.
We also allow kernel functions $k_{ql}$ and bandwidths $h_{ql}$
to depend on the state $q$ and lag $l$ considered. In addition to
$\boldsymbol{K}$, we also have standard HMM parameters describing
the hidden-state evolution, specifically a matrix $\boldsymbol{A}\in\mathbb{R}^{M\times M}$
of state transition probabilities defined by 
\begin{align}
a_{qq'} & =P\left(Q_{t+1}=q'\mid Q_{t}=q\right)\textrm{.}
\end{align}
 (Since the model is stationary and ergodic, the leading left eigenvector
$\boldsymbol{\pi}\in\mathbb{R}^{M}$ of $\boldsymbol{A}$ satisfying
$\sum_{q}\pi_{q}=1$ defines the stationary state distribution $P\left(Q_{t}=q\right)=\pi_{q}$
of the Markov chain. This also gives the initial state probabilities
of the model.) The full set of KDE-HMM parameters is thus $\boldsymbol{\theta}=\left\{ \boldsymbol{A},\,\boldsymbol{K}\right\} $.
In Section \ref{paperd:sec:training} we derive update formulas to
estimate these parameters from data.

To the best of our knowledge, models of the above form have not been
considered previously. KDE-HMMs generalize \cite{rajarshi1990a} and
\cite{manzana2008a} by introducing hidden states (when $M>1$) and
different kernels and bandwidths for every lag $l$. Our proposal
is also more general than HMMs with KDE outputs, dubbed KDE/HMM, previously
investigated in \cite{piccardi2007a}. We recover KDE/HMMs by setting
$p=0$, making samples conditionally independent given the state sequence,
similar to Fig.\ \ref{paperd:fig:hmmdeps}.

\begin{figure}[tp]
\begin{centering}
\includegraphics[width=0.48\columnwidth]{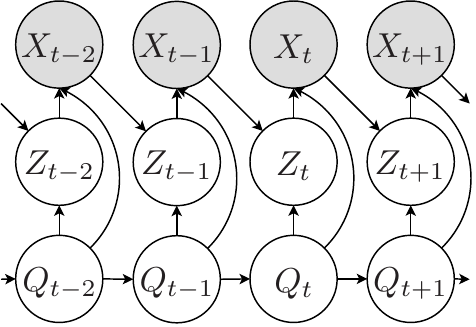}
\par\end{centering}
\caption{\label{paperd:fig:kde1-hmmdeps}Variable dependences in a first-order
KDE-HMM. The state $Q$ and component $Z$ are unobserved latent variables.}
\vspace{-2ex}
\end{figure}

Like with KDE-MMs, the next-step distribution of KDE-HMMs can be broken
into two parts: one, $\kappa_{qn}$, choosing an exemplar $Z_{t}\in\left\{ p+1,\,\ldots,\,N\right\} $
from $\mathcal{D}$ for the next step, given the context $\underline{x}_{t-p}^{t-1}$
and the current state $q_{t}$, and the other adding kernel-shaped
noise around the selected exemplar value $y_{z_{t}}$. The dependences
among variables in a KDE-HMM with $p=1$ are illustrated in Fig.\ \ref{paperd:fig:kde1-hmmdeps}.
Despite the increased complexity, sampling is straightforward, and
efficient inference remains possible for all $p$ using standard forward-backward
recursions. Specifically, past and future values of $Q$ and $Z$
are conditionally independent given the sequence $\underline{x}$
and the state $q_{t}$ at time $t$, as can be verified, e.g., using
the algorithm in \cite{shachter1998a}.

Adding a hidden state has several advantages. Aside from the benefits
discussed in Sections \ref{paperd:sub:hmms} and \ref{paperd:sub:combinationmodels},
the state essentially partitions the space of contexts $\underline{x}_{t-p}^{t-1}$
into regions associated with different models, thus allowing different
bandwidths for different situations. This can be a significant advantage
in many cases\textemdash even when the data-generating process is
Markovian\textemdash since it often is desirable to apply different
degrees of smoothing in the modes versus the tails of distributions:
see, for instance, \cite{silverman1986a} or \cite{marchette1994a},
where the latter's filtering functions $\rho_{j}\left(\cdot\right)$
closely resemble our weights $w_{qn}$. Letting the bandwidth depend
on the lag $l$ furthermore allows better representation of phenomena
such as variable (typically decreasing) predictive value of older
samples, and separates context sensitivity ($h_{ql}$ for $l>0$)
from the added noise ($h_{ql}$ for $l=0$) during sampling. This
distinguishes KDE-HMMs from KDE-MMs even if the number of states,
$M$, is one. As before, the advantages come at a price of greater
demands for data and computation, compared to parametric approaches.

\section{Parameter Estimation\label{paperd:sec:training}}

To use models in practice, we must identify parameters that fit well
for a given process of interest. In this section, we consider bandwidth
selection and parameter estimation for KDE-MMs and KDE-HMMs. In particular,
we derive guaranteed-ascent maximum likelihood update equations based
on the EM-algorithm for the case of Gaussian kernels, along with relaxed
update formulas which converge faster in practice.

\subsection{Parameter Estimation Objective Function}

Numerous principles have been proposed for choosing bandwidths in
a data-driven manner (see \cite{cau1994a} and \cite{turlach1993a}
for reviews). In practice, the preferred scheme depends on the data
and on the intended application, including the performance measure
(loss function) of interest. In this paper we concentrate on the classic
Kullback-Leibler (or KL) divergence from information theory, 
\begin{align}
D_{\mathrm{KL}}\left(f_{\boldsymbol{X}}\,\middle|\middle|\,\widehat{f}_{\boldsymbol{X}}\right) & =\int f_{\boldsymbol{X}}\left(\boldsymbol{x}\right)\ln\frac{f_{\boldsymbol{X}}\left(\boldsymbol{x}\right)}{\widehat{f}_{\boldsymbol{X}}\left(\boldsymbol{x}\right)}\mathrm{d}\boldsymbol{x}\textrm{,}\label{paperd:eq:kldiv}
\end{align}
 previously used with KDE in signal processing and machine learning
by \cite{piccardi2007a} and \cite{holmes2007a}. Minimizing $D_{\mathrm{KL}}$
between empirical and estimated distributions leads to traditional
maximum-likelihood (ML) parameter estimation, cf.\ \cite{akaike1973a}.
While likelihood maximization is not considered appropriate for heavy-tailed
reference densities $f_{\boldsymbol{X}}$ \cite{broniatowski1989a,cau1994a},
likelihoods are asymptotically a factor $N$ faster to evaluate compared
to other bandwidth-selection criteria such as integrated squared error
\cite{holmes2007a}.

The likelihood function is given by Eqs.\ (\ref{paperd:eq:kde-mm})
and (\ref{paperd:eq:kde-hmm}) applied to the training data sequence
$\mathcal{D}=\underline{y}_{1}^{N}$, but it is not appropriate to
optimize this function directly. In the limit where bandwidth goes
to zero, the KDE $\widehat{f}_{\boldsymbol{X}}$ shrinks to a set
of point masses (spikes) placed at the points in $\mathcal{D}$ and
the likelihood diverges to positive infinity. A similar issue exists
in common models such as Gaussian mixture models (GMMs) and Gaussian-output
HMMs: by dedicating one component or state to a narrow spike centred
on a single datapoint, arbitrarily high likelihood values can be achieved
\cite[pp.\ 433--434]{bishop2006a}.

With KDE, a standard workaround is to use cross-validation, for every
$t$ omitting the component centred at $y_{t}$ when computing the
likelihood term at $\widehat{f}_{X}\left(y_{t};\,h\right)$. This
prevents points in the training data from ``explaining themselves''
and removes the degenerate optimum at zero. The resulting objective
function $\widetilde{f}$ is known as the \emph{pseudo-likelihood}.
It can be shown that, under certain conditions, maximizing the pseudo-likelihood
asymptotically minimizes the KL-divergence of KDE \cite{hall1987a}.

For simplicity, we exclude the first $p$ points of all sequences,
where the context is incomplete, from the KDE-HMM likelihood evaluations
in this paper. (KDE-MMs use the periodic extension from footnote \ref{paperd:fn:kde-mmperiodic}.)
Using (\ref{paperd:eq:kde-hmmweights}) and (\ref{paperd:eq:kde-hmm})
then leads to the pseudo-likelihood 
\begin{gather}
\widetilde{f}_{\underline{X}}\left(\underline{y}_{1}^{N};\,\boldsymbol{\theta}\right)=\sum_{\underline{q}_{p+1}^{N}}\sum_{\underline{z}_{p+1}^{N}}\widetilde{f}_{\underline{X},\,\underline{Q},\,\underline{Z}}\left(\underline{y}_{1}^{N},\,\underline{q}_{p+1}^{N},\,\underline{z}_{p+1}^{N};\,\boldsymbol{\theta}\right)\label{paperd:eq:pseudolike}
\end{gather}
 \vspace{-14pt}
\begin{multline}
\widetilde{f}_{\underline{X},\,\underline{Q},\,\underline{Z}}\left(\underline{y}_{1}^{N},\,\underline{q}_{p+1}^{N},\,\underline{z}_{p+1}^{N};\,\boldsymbol{\theta}\right)=P\left(\underline{Q}_{p+1}^{N}=\underline{q}_{p+1}^{N};\,\boldsymbol{A}\right)\\
\cdot\prod_{t=p+1}^{N}\frac{\left(1-\delta_{z_{t}t}\right)w_{q_{t}z_{t}}\prod_{l=0}^{p}k_{q_{t}l}\left(\frac{y_{t-l}-y_{z_{t}-l}}{h_{q_{t}l}}\right)}{h_{q_{t}0}\sum_{n=p+1,\,n\neq t}^{N}w_{q_{t}n}\prod_{l=1}^{p}k_{q_{t}l}\left(\frac{y_{t-l}-y_{n-l}}{h_{q_{t}l}}\right)}\textrm{.}\label{paperd:eq:yzpseudolike}
\end{multline}
 The cross-validation terms $\left(1-\delta_{z_{t}t}\right)$ are
introduced to set the probabilities of sequences having $z_{t}=t$
to zero.

To reduce clutter, we will from now on take the indices in expressions
and summations to range as $q\in\left\{ 1,\,\ldots,\,M\right\} $,
$l\in\left\{ 1,\,\ldots,\,p\right\} $, and $t,\,n\in\left\{ p+1,\,\ldots,\,N\right\} $,
unless otherwise specified. Primed indices follow the same limits
as unprimed ones.

\subsection{Expectation Maximization for KDE-HMMs}

As with other latent-variable models such as GMMs and HMMs, direct
analytic optimization of the (log) pseudo-likelihood is infeasible
due to the sums over latent variables in (\ref{paperd:eq:pseudolike}).
Instead, we seek an iterative optimization procedure based on the
expectation maximization (EM) algorithm \cite{dempster1977a}, by
maximizing the \emph{auxiliary function} 
\begin{multline}
\mathcal{Q}\left(\boldsymbol{\theta}^{\prime};\,\widehat{\boldsymbol{\theta}}\right)\\
=\mathbb{E}\left(\ln\widetilde{f}_{\underline{X},\,\underline{Q},\,\underline{Z}}\left(\underline{y},\,\underline{Q},\,\underline{Z};\,\boldsymbol{\theta}^{\prime}\right)\mid\underline{X}=\underline{y};\,\widehat{\boldsymbol{\theta}}\right)\textrm{.}\label{paperd:eq:emauxiliary}
\end{multline}

Using Jensen's inequality, one can prove that any revised parameter
estimate $\widehat{\boldsymbol{\theta}}^{\left(\mathrm{new}\right)}$
satisfying 
\begin{align}
\mathcal{Q}\Big(\widehat{\boldsymbol{\theta}}^{\left(\mathrm{new}\right)};\,\widehat{\boldsymbol{\theta}}\Big) & \geq\mathcal{Q}\left(\widehat{\boldsymbol{\theta}};\,\widehat{\boldsymbol{\theta}}\right)\label{paperd:eq:auxincrease}
\end{align}
 is guaranteed not to decrease the actual likelihood of the data;
this is used to establish convergence. In particular, convergence
does not require maximizing (\ref{paperd:eq:emauxiliary}) at every
step, and procedures that increase $\mathcal{Q}$ without necessarily
maximizing it are termed \emph{generalized EM} (GEM). Because the
pseudo-likelihood is equivalent to the regular likelihood together
with a specific cross-validation prior on $Z$, convergence is assured
also in our case.

At each EM iteration the shape of the auxiliary function is determined
by the conditional posterior distribution of the latent variables
$\underline{Q}$ and $\underline{Z}$. For KDE-HMMs, the relevant
forward-backward recursions to determine the conditional hidden-state
distributions, also known as \emph{state occupancies}, 
\begin{align}
\gamma_{qt} & =P\left(Q_{t}=q\mid\underline{X}_{1}^{N}=\underline{y}_{1}^{N};\,\widehat{\boldsymbol{\theta}}\right)\label{paperd:eq:emoccupancies}
\end{align}
 are identical to those of other HMMs with Markovian output distributions,
such as \cite{shannon2009a}, apart from enforcing $P\left(Z_{t}=t\right)=0\,\forall t$
due to cross-validation. The occupancies are used to update the state
transition probabilities $\boldsymbol{A}$ of the KDE-HMM following
standard HMM formulas available in \cite{rabiner1989a}.

Determining conditional posterior distributions, or \emph{responsibilities},
of the latent KDE component variables $Z_{t}$ is similarly straightforward,
and one obtains 
\begin{align}
\varrho_{qnt}^{\mathrm{num}} & =P\left(Z_{t}=n\mid Q_{t}=q,\,\underline{X}_{t-p}^{t}=\underline{y}_{t-p}^{t};\,\widehat{\boldsymbol{K}}\right)\\
 & =\frac{\left(1-\delta_{nt}\right)\kappa_{qn}\left(\underline{y}_{t-p}^{t-1};\,\widehat{\boldsymbol{K}}\right)k_{q0}\left(\frac{y_{t}-y_{n}}{\widehat{h}_{q0}}\right)}{\sum_{n'\neq t}\kappa_{qn'}\left(\underline{y}_{t-p}^{t-1};\,\widehat{\boldsymbol{K}}\right)k_{q0}\left(\frac{y_{t}-y_{n'}}{\widehat{h}_{q0}}\right)}\textrm{.}\label{paperd:eq:numrhos}
\end{align}
 The auxiliary function for weights and bandwidth parameters $\boldsymbol{K}$
then takes the form 
\begin{multline}
\mathcal{Q}_{\boldsymbol{K}}\left(\boldsymbol{K}^{\prime};\,\widehat{\boldsymbol{K}}\right)=\sum_{q,\,t,\,n\neq t}\gamma_{qt}\varrho_{qnt}^{\mathrm{num}}\left(\ln w_{qn}^{\prime}-\frac{1}{2}\ln h_{q0}^{\prime}\right)\\
+\sum_{q,\,t,\,n\neq t}\gamma_{qt}\varrho_{qnt}^{\mathrm{num}}\sum_{l=0}^{p}\ln k_{ql}\left(\frac{y_{t-l}-y_{n-l}}{h_{ql}^{\prime}}\right)\\
-\sum_{q,\,t}\gamma_{qt}\ln\left(\sum_{n\neq t}w_{qn}^{\prime}\prod_{l}k_{ql}\left(\frac{y_{t-l}-y_{n-l}}{h_{ql}^{\prime}}\right)\right)\textrm{.}\label{paperd:eq:kdeauxiliary}
\end{multline}
 Optimal values for the next-step bandwidths $h_{q0}^{\prime}$ are
readily identified \cite{piccardi2007a} and are presented in (\ref{paperd:eq:nextstephupdate})
below. The weights and context bandwidths ($w_{qn}^{\prime}$ and
$h_{ql}^{\prime}$ for $l>0$) present an obstacle, however, as these
appear in a negative term with a sum inside a logarithm and cannot
be optimized analytically. Such terms are common in models that feature
conditional probabilities with renormalization, e.g., standard maximum
mutual information (MMI) discriminative classifiers \cite{bahl1986a}.

To identify GEM updates of $\boldsymbol{K}$ with Gaussian kernels
we construct a global lower bound $\underline{\mathcal{Q}}_{\boldsymbol{K}}$
of $\mathcal{Q}_{\boldsymbol{K}}$ in (\ref{paperd:eq:kdeauxiliary})
which is tight at $\boldsymbol{K}^{\prime}=\widehat{\boldsymbol{K}}$.
Any updated parameters $\boldsymbol{K}^{\prime}$ that does not decrease
$\underline{\mathcal{Q}}_{\boldsymbol{K}}$ is then guaranteed not
to decrease the true auxiliary function, and the key convergence relation
(\ref{paperd:eq:auxincrease}) remains satisfied. This approach is
known as \emph{minorize-maximization} and can be seen as a generalization
of EM \cite{hunter2004a}.

While many bounds on log-sum-exp-type expressions exist \cite{bouchard2007a},
we chose to base our bound $\underline{\mathcal{Q}}_{\boldsymbol{K}}$
on so-called \emph{reverse-Jensen bounds} \cite{jebara2000a,jebara2001a}.
These apply to Gaussian and other exponential-family kernels that
satisfy the conditions of Lemma~2, page~139 in \cite{jebara2001a}.
Importantly, the reverse-Jensen bounds have the same parametric dependence
on $\boldsymbol{K}^{\prime}$ as the other terms in Eq.\ (\ref{paperd:eq:kdeauxiliary}),
enabling us to to derive analytic expressions for the parameters $\widehat{\boldsymbol{K}}^{\left(\mathrm{new}\right)}$
that maximize the lower bound $\underline{\mathcal{Q}}_{\boldsymbol{K}}$.

A mathematically similar application of reverse-Jensen bounds to MMI
training of HMM classifiers can be found in \cite{afify2005a}. The
resulting updates resemble those produced by a common MMI training
technique called \emph{extended Baum-Welch} (EBW) \cite{gopalakrishnan1991a,normandin1994a}.
However, while standard EBW equations include a heuristically set
\cite{woodland2002a} tuning parameter that limits the update magnitude,
the reverse-Jensen procedure selects the tuning parameter automatically,
requiring no user intervention.

\subsection{Generalized EM Update Formulas for KDE-HMM}

For Gaussian kernels as in (\ref{paperd:eq:gausskern}), maximizing
the reverse-Jensen lower bound yields the GEM update equations 
\begin{align}
\widehat{h}_{q0}^{2\left(\mathrm{new}\right)} & =\dfrac{\sum_{t,\,n\neq t}\gamma_{qt}\varrho_{qnt}^{\mathrm{num}}\left(y_{t}-y_{n}\right)^{2}}{\sum_{t,\,n\neq t}\gamma_{qt}\varrho_{qnt}^{\mathrm{num}}}\label{paperd:eq:nextstephupdate}\\
\widehat{h}_{ql}^{2\left(\mathrm{new}\right)} & =\dfrac{W_{q}\widehat{h}_{ql}^{2}+\sum_{t,\,n\neq t}\gamma_{qnt}^{\mathrm{diff}}\left(y_{t-l}-y_{n-l}\right)^{2}}{W_{q}+\sum_{t,\,n\neq t}\gamma_{qnt}^{\mathrm{diff}}}\label{paperd:eq:contexthupdate}\\
\widehat{w}_{qn}^{\left(\mathrm{new}\right)} & =\frac{W_{q}\widehat{w}_{qn}+\sum_{t}\gamma_{qnt}^{\mathrm{diff}}}{W_{q}+\sum_{t,\,n'\neq t}\gamma_{qn't}^{\mathrm{diff}}}\label{paperd:eq:wupdate}\\
\gamma_{qnt}^{\mathrm{diff}} & =\gamma_{qt}\left(\varrho_{qnt}^{\mathrm{num}}-\varrho_{qnt}^{\mathrm{den}}\right)\label{paperd:eq:gammadiff}\\
W_{q} & =\sum_{t,\,n\neq t}\gamma_{qt}\left(\varrho_{qnt}^{\mathrm{den}}+\omega_{qnt}^{h}+\omega_{qnt}^{w}+\omega_{qnt}^{\prime}\right)\textrm{;}
\end{align}
 the state occupancies $\gamma_{qt}$ in these formulas are defined
by (\ref{paperd:eq:emoccupancies}), while the responsibilities $\varrho_{qnt}^{\mathrm{num}}$
and the terms from the reverse-Jensen bound follow (\ref{paperd:eq:numrhos})
and 
\begin{align}
\varrho_{qnt}^{\mathrm{den}} & =P\left(Z_{t}=n\mid Q_{t}=q,\,\underline{X}_{t-p}^{t-1}=\underline{y}_{t-p}^{t-1};\,\widehat{\boldsymbol{K}}\right)\\
 & =\frac{\left(1-\delta_{nt}\right)\kappa_{qn}\left(\underline{y}_{t-p}^{t-1};\,\widehat{\boldsymbol{K}}\right)}{\sum_{n'\neq t}\kappa_{qn'}\left(\underline{y}_{t-p}^{t-1};\,\widehat{\boldsymbol{K}}\right)}\label{paperd:eq:denrhos}\\
\omega_{qnt}^{h} & =2G\left(\frac{1}{2}\varrho_{qnt}^{\mathrm{den}}\right)\sum_{l=1}^{p}\left(\xi_{ql}^{h}\right)^{2}\label{paperd:eq:omegahoriginal}\\
\omega_{qnt}^{w} & =4G\left(\frac{1}{2}\varrho_{qnt}^{\mathrm{den}}\right)\xi_{qn}^{w}\\
\omega_{qnt}^{\prime} & =\varrho_{qnt}^{\mathrm{den}}\max\left(\max_{l}\left(\xi_{ql}^{h}\right),\,\xi_{qn}^{w}\right)\label{paperd:eq:omegap}\\
\xi_{ql}^{h} & =\left(y_{t-l}-y_{n-l}\right)^{2}\widehat{h}_{ql}^{-2}-1\label{paperd:eq:xih}\\
\xi_{qn}^{w} & =\widehat{w}_{qn}^{-1}-1\textrm{.}
\end{align}
 The function 
\begin{align}
G\left(\gamma\right) & =\begin{cases}
\left(\frac{\gamma-1}{\ln\gamma}\right)^{2}-\frac{1}{4\ln\gamma} & \textrm{if }\gamma<\nicefrac{1}{6}\\
\left(\frac{\nicefrac{1}{6}-1}{\ln\nicefrac{1}{6}}\right)^{2}-\frac{1}{4\ln\nicefrac{1}{6}}+\gamma-\frac{1}{6} & \textrm{otherwise}
\end{cases}
\end{align}
 follows Eq.\ (5.10) in \cite[p.\ 99]{jebara2001a}.

The update formulas in (\ref{paperd:eq:nextstephupdate}) through
(\ref{paperd:eq:wupdate}) should increase the pseudo-likelihood except
at local extreme points. Since 
\begin{align}
W_{q}+\sum_{t,\,n\neq t}\gamma_{qnt}^{\mathrm{diff}} & >0\textrm{,}
\end{align}
 the updated parameters are always well defined.
\begin{prop}
Let $\widehat{\boldsymbol{\theta}}^{\left(i\right)}$ for $i\in\mathbb{Z}_{+}$
be a sequence of parameter estimates, computed iteratively from $\mathcal{D}=\underline{y}_{1}^{N}$
using the update procedure above, and let $\widetilde{f}^{\left(i\right)}$
be the corresponding pseudo-likelihoods in Eq.\ (\ref{paperd:eq:pseudolike}).
Define the minimum separation $d_{\mathrm{min}}$ through 
\begin{align}
d_{\mathrm{min}} & =\min_{t,\,n\neq t}\left|y_{t}-y_{n}\right|\textrm{.}
\end{align}
 Assume $d_{\mathrm{min}}$ is strictly positive and $N>p+1$. Then
$\lim_{i\to\infty}\widetilde{f}^{\left(i\right)}$ exists.
\end{prop}
The proof proceeds by establishing that the pseudo-likelihood has
a finite upper bound $\widetilde{f}_{\mathrm{max}}$, and then showing
that $\widetilde{f}_{\mathrm{max}}\geq\widetilde{f}^{\left(i+1\right)}\geq\widetilde{f}^{\left(i\right)}\,\forall i$,
which ensures convergence.

\subsection{Accelerated Updates\label{paperd:sub:fastkde-hmmupdates}}

Studying the update equations, we see that the quantity $W_{q}$ limits
the update step length: if $W_{q}$ is large compared to $\left|\sum_{t,\,n\neq t}\gamma_{qnt}^{\mathrm{diff}}\right|$,
we have $\widehat{h}_{ql}^{\left(\mathrm{new}\right)}\approx\widehat{h}_{ql}$,
and similarly for $\widehat{w}_{qn}^{\left(\mathrm{new}\right)}$.
Unfortunately, $W_{q}$ often substantially exceeds $\left|\sum_{t,\,n\neq t}\gamma_{qnt}^{\mathrm{diff}}\right|$
on large datasets, and the time required until eventual convergence
is then infeasibly long.

To reduce $W_{q}$ and obtain formulas that produce larger updates,
one may let the weights $w_{qn}$ be fixed (as their updates drive
up the $W_{q}$-factors the most) and only consider updating the bandwidths.
We can then set $\xi_{qn}^{w}=0$ in all formulas. Moreover, we apply
the approximation $G\left(\gamma\right)\approx\gamma$, which is the
first of several steps involved in connecting reverse-Jensen update
formulas to standard EBW heuristics \cite{afify2005a}. This yields
the approximate weights 
\begin{align}
\widetilde{\omega}_{qnt}^{h} & =\varrho_{qnt}^{\mathrm{den}}\sum_{l}\left(\xi_{ql}^{h}\right)^{2}\textrm{,}\label{paperd:eq:relaxedomegah}
\end{align}
and the associated relaxed $W_{q}$-factors 
\begin{align}
\widetilde{W}_{q} & =\sum_{t,\,n\neq t}\gamma_{qt}\left(\varrho_{qnt}^{\mathrm{den}}+\widetilde{\omega}_{qnt}^{h}+\omega_{qnt}^{\prime}\right)\textrm{.}\label{paperd:eq:relaxedW}
\end{align}
 (It is important to remember that $\omega_{qnt}^{\prime}$ here should
use $\xi_{qn}^{w}=0$.) The resulting KDE-HMM training algorithm is
summarized in Table \ref{paperd:tab:trainingalgo}.
\begin{table}[tp]
\caption{\label{paperd:tab:trainingalgo}Accelerated training of Guassian-kernel
KDE-HMMs.}
\vspace{-1ex}
\begin{footnotesize}
\begin{algorithmic}[0]
\Function{TrainKDE-HMM}{$\underline{y}_{1}^{N},\,\widehat{\boldsymbol{A}},\,w_{qn},\,\widehat{h}_{q0},\,\widehat{h}_{ql}$}
\State $\xi_{qn}^{w} \leftarrow 0$
\State compute initial state distribution $\widehat{\boldsymbol{\pi}}$ from $\widehat{\boldsymbol{A}}$
\Repeat
\State compute $\widehat{\alpha}_{qt},\,\widehat{\beta}_{qt},\,\gamma_{qt}$ using forward-backward \cite{rabiner1989a}
\State re-estimate $\widehat{\boldsymbol{A}}$ using standard formula \cite{rabiner1989a}
\State recompute $\widehat{\boldsymbol{\pi}}$ from $\widehat{\boldsymbol{A}}$
\State compute $\varrho_{nt}^{\mathrm{num}},\,\varrho_{nt}^{\mathrm{den}}$ using \eqref{paperd:eq:numrhos}, \eqref{paperd:eq:denrhos}
\State compute $\gamma_{qnt}^{\mathrm{diff}}$ using \eqref{paperd:eq:gammadiff}
\State compute $\xi_{ql}^{h}$ using \eqref{paperd:eq:xih}
\State compute $\widetilde{\omega}_{qnt}^{h},\,\omega_{qnt}^{\prime}$ using \eqref{paperd:eq:relaxedomegah}, \eqref{paperd:eq:omegap}
\State compute $\widetilde{W}_{q}$ using \eqref{paperd:eq:relaxedW}
\State re-estimate $\widehat{h}_{q0},\,\widehat{h}_{ql}$ using \eqref{paperd:eq:nextstephupdate}, \eqref{paperd:eq:contexthupdate}
\Until{convergence}
\State \Return $\widehat{\boldsymbol{A}},\,w_{qn},\,\widehat{h}_{q0},\,\widehat{h}_{ql}$
\EndFunction
\end{algorithmic}
\end{footnotesize}

\vspace{-2ex}
\end{table}

Since $\widetilde{W}_{q}$ in (\ref{paperd:eq:relaxedW}) tends to
be of roughly the same order of magnitude as $\left|\sum_{t,\,n\neq t}\gamma_{qnt}^{\mathrm{diff}}\right|$,
EM-training with $\widetilde{W}_{q}$ instead of $W_{q}$ converges
quite quickly. On the other hand, $\widetilde{W}_{q}$ remains sufficiently
conservative to virtually always increase the likelihood at every
step in our experiments.

\subsection{KDE-MM Bandwidth Selection}

We now turn to consider the special case of KDE-MMs. While bandwidth
selection formulas for KDE-MM-like models do exist in the literature,
e.g., \cite{paparoditis2002a}, these focus on mean squared error
rather than pseudo-likelihood. Since KDE-MMs essentially are single-state
KDE-HMMs with fixed, uniform weights $w_{1n}=\frac{1}{N}$ and all
bandwidths tied to be equal, the approach in this paper can also be
used to derive pseudo-likelihood maximization formulas for the KDE-MM
bandwidth $h$. Assuming a Gaussian kernel this yields the iterative
updates 

\begin{align}
\widehat{h}^{2\left(\mathrm{new}\right)} & =\dfrac{W\widehat{h}^{2}+\sum_{t,\,n\neq t}\left(\varrho_{nt}^{\mathrm{num}}\left(y_{t}-y_{n}\right)^{2}+p\varrho_{nt}^{\mathrm{diff}}\overline{d}_{nt}^{2}\right)}{W+\sum_{t,\,n\neq t}\left(\varrho_{nt}^{\mathrm{num}}+p\varrho_{nt}^{\mathrm{diff}}\right)}\label{paperd:eq:kdemmhupdate}\\
\varrho_{nt}^{\mathrm{diff}} & =\varrho_{nt}^{\mathrm{num}}-\varrho_{nt}^{\mathrm{den}}\\
W & =p\sum_{t,\,n\neq t}\left(\varrho_{nt}^{\mathrm{den}}+\omega_{nt}+\omega_{nt}^{\prime}\right)\\
\omega_{nt} & =2pG\left(\frac{1}{2}\varrho_{nt}^{\mathrm{den}}\right)\left(\overline{d}_{nt}^{2}\widehat{h}^{-2}-1\right)^{2}\\
\omega_{nt}^{\prime} & =\varrho_{nt}^{\mathrm{den}}\max\left(\overline{d}_{nt}^{2}\widehat{h}^{-2}-1,\,0\right)\\
\overline{d}_{nt}^{2} & =\frac{1}{p}\sum_{l}\left(y_{t-l}-y_{n-l}\right)^{2}\textrm{.}
\end{align}
 with $\varrho_{nt}^{\mathrm{num}}$ and $\varrho_{nt}^{\mathrm{den}}$
defined as in Eqs.\ (\ref{paperd:eq:numrhos}) and (\ref{paperd:eq:denrhos}),
respectively, but with all references to the state $q$ omitted. Just
as for KDE-HMMs, the approximation $G\left(\gamma\right)\approx\gamma$
can be introduced to obtain relaxed weights 
\begin{align}
\widetilde{W} & =p\sum_{t,\,n\neq t}\left(\varrho_{nt}^{\mathrm{den}}+\widetilde{\omega}_{nt}+\omega_{nt}^{\prime}\right)\\
\widetilde{\omega}_{nt} & =p\varrho_{nt}^{\mathrm{den}}\left(\overline{d}_{nt}^{2}\widehat{h}^{-2}-1\right)^{2}
\end{align}
 which increase the stepsize of the updates. Alternatively, it is
straightforward to maximize the pseudo-likelihood using general-purpose
numerical optimization methods, as KDE-MMs only have a single free
parameter.

\subsection{Initialization\label{paperd:sub:initialization}}

Similar to traditional training schemes for HMMs, the proposed parameter
estimation schemes for KDE-MMs and KDE-HMMs rely on iterative refinements,
and thus require initialization. For standard EM-training of models
such as HMMs and AR-HMMs, it is sufficient to provide an initial guess
$\widehat{\gamma}_{qt}$ of the state occupancies given $\underline{y}_{1}^{N}$
to begin estimating parameters. In contrast, our KDE-HMM update formulas,
like EBW, depend on previous parameter values. Initial weights and
bandwidth parameters must therefore be assigned explicitly. We propose
to set the weight parameters based on state occupancies according
to 
\begin{align}
\widehat{w}_{qn} & =\frac{\widehat{\gamma}_{qn}}{\sum_{n'}\widehat{\gamma}_{qn'}}\textrm{,}\label{paperd:eq:initialweights}
\end{align}
 as an estimate of the conditional component probabilities $P\left(Z_{t}=n\mid Q_{t}=q;\,\boldsymbol{\theta}\right)$.
Initial bandwidths $\widehat{h}_{q0}$ and $\widehat{h}_{ql}$ can
then be set using the multidimensional KDE ``normal reference rule''
from \cite{scott1992a,bowman1997a}, with each point weighted according
to $w_{qn}$, while the transition matrix may be initialized based
on co-occurrences, 
\begin{align}
\widehat{a}_{qq'} & =\frac{\sum_{t=1}^{N-1}\widehat{\gamma}_{qt}\widehat{\gamma}_{q'\left(t+1\right)}}{\sum_{t=1}^{N-1}\widehat{\gamma}_{qt}}\textrm{.}\label{paperd:eq:initialtrans}
\end{align}

Sometimes occupancy estimates $\widehat{\gamma}_{qn}$ for seeding
the model can be computed based on domain knowledge or by inspecting
the data. This approach will be used for the experiments in this article.
Another common principle is to base advanced models on faster, simpler
methods, e.g., using $k$-means to initialize GMMs. For KDE-HMMs it
is natural to set $\widehat{\gamma}_{qn}=\gamma_{qn}^{\left(\mathrm{HMM}\right)}$,
where $\gamma_{qn}^{\left(\mathrm{HMM}\right)}$ are the state occupancies
of a trained parametric hidden-state model such as an HMM or an AR-HMM.

The mechanics of KDE-HMM initialization are summarized in Table \ref{paperd:tab:initalgos}.
Given a trained model, the probability of any given observation sequence
under $\boldsymbol{\theta}$ can be evaluated with the forward algorithm,
just like for regular HMMs \cite{rabiner1989a}. Sequentially generating
samples from the model is similarly straightforward.
\begin{table}[tp]
\caption{\label{paperd:tab:initalgos}Initialization of Gaussian-kernel KDE-HMMs.}
\vspace{-1ex}
\begin{footnotesize}
\begin{algorithmic}[0]
\Function{GammasToKDE-HMM}{$\underline{y}_{1}^{N},\,\widehat{\gamma}_{qt}$}
\State compute $\widehat{\boldsymbol{A}}$ using \eqref{paperd:eq:initialtrans}
\State compute $\widehat{w}_{qn}$ using \eqref{paperd:eq:initialweights}
\State compute $\widehat{h}_{q0},\,\widehat{h}_{ql}$ using weighted reference rule \cite{scott1992a,bowman1997a}
\State \Return $\widehat{\boldsymbol{A}},\,\widehat{w}_{qn},\,\widehat{h}_{q0},\,\widehat{h}_{ql}$
\EndFunction
\end{algorithmic}

\begin{algorithmic}[0]
\Function{HMMToKDE-HMM}{$\underline{y}_{1}^{N}$, an HMM $\boldsymbol{\lambda}$}
\State compute $\widehat{\gamma}_{qt}$ from $\boldsymbol{\lambda}$ using forward-backward \cite{rabiner1989a}
\State take $\widehat{\boldsymbol{A}}$ from $\boldsymbol{\lambda}$
\State compute $\widehat{w}_{qn}$ using \eqref{paperd:eq:initialweights}
\State compute $\widehat{h}_{q0},\,\widehat{h}_{ql}$ using weighted reference rule \cite{scott1992a,bowman1997a}
\State \Return $\widehat{\boldsymbol{A}},\,\widehat{w}_{qn},\,\widehat{h}_{q0},\,\widehat{h}_{ql}$
\EndFunction
\end{algorithmic}
\end{footnotesize}

\vspace{-2ex}
\end{table}

\section{Experiments on Synthetic Data\label{paperd:sec:synthexperiments}}

To investigate the probabilistic modelling capabilities of KDE-MMs
and KDE-HMMs, we applied these techniques to a selection of datasets,
and compared their prediction performance against other, standard
time-series models representing the different time-dependence paradigms
discussed in Section \ref{paperd:sec:background}.

In this section, we consider an application to synthetic data, illustrating
the strong, general asymptotic convergence properties of KDE-based
models, also for non-Gaussian processes that baseline predictors cannot
describe. Applications to nonlinear real-life datasets are considered
in Section \ref{paperd:sec:realexperiments}.

\subsection{Data Series}

As a first test, we generated data from a simple reference process
having both hidden-state and Markovian dependences as in Fig.\ \ref{paperd:fig:ar2-hmmdeps}.
Specifically, we used a first-order linear AR-HMM with two states
as the data source. We let both state-conditional AR-processes have
the same mean (zero) and correlation coefficient $\nicefrac{2}{3}$,
but gave one state a much larger standard deviation for the driving
noise ($\sigma_{1}=1$ versus $\sigma_{2}=5$). We can write this
process as 
\begin{align}
X_{t} & =\frac{2}{3}X_{t-1}+\sigma_{Q_{t}}U_{t}\textrm{,}
\end{align}
 where $U_{t}$ is zero-mean, unit-variance white noise. A symmetric
hidden-state transition matrix $\boldsymbol{A}$ was chosen, with
a probability $a_{qq}=\nicefrac{4}{5}$ of staying in the same state,
and $1-a_{qq}=\nicefrac{1}{5}$ of switching to the other state at
each time step. This produced time series cycling through volatile
and quiescent periods, similar to stock market data.

We considered two variations of the above model, differing in the
properties of the driving noise $U_{t}$. In the first application,
we let $U_{t}$ be i.i.d.\ standard normal distributed, while in
the second, we used independent samples from the bimodal zero-mean
Gaussian mixture 
\begin{gather}
f_{U_{t}}\left(u\right)=\frac{1}{\sqrt{2\pi}\sigma_{U}}\sum_{c=0}^{1}\exp\left(-\frac{1}{2}\left(\frac{u-\left(-1\right)^{c}\mu_{U}}{\sigma_{U}}\right)^{2}\right)\\
\mu_{U}=\sqrt{\frac{36}{37}}\textrm{,}\qquad\sigma_{U}=\frac{1}{\sqrt{37}}\textrm{,}
\end{gather}
 which also has unit variance.

For each of the two data sources above, five different training data
sizes $N$ between $N=10^{1.5}$ and $N=10^{3.5}$ were considered.
For each data size, 30 independent realizations were generated of
each process, along with 30 independent validation sets of $T=1000$
samples.

\subsection{Experiment Setup\label{paperd:sub:synthsetup}}

We applied five different models to the datasets to investigate model
convergence speed and asymptotic prediction performance as a function
of training data size. The tested models were a first-order linear
Gaussian autoregressive model; a two-state Gaussian-output HMM; a
two-state, first-order Gaussian AR-HMM (combining the first two models);
a first-order KDE-MM trained by numerically optimizing the pseudo-likelihood;
and a two-state, first-order KDE-HMM trained using the algorithm in
Table \ref{paperd:tab:trainingalgo}. The models are summarized in
Table \ref{paperd:tab:models}.

All hidden-state models in this paper, KDE-HMMs included, were initialized
based on estimated state occupancies $\widehat{\gamma}_{qn}$ as outlined
in Section \ref{paperd:sub:initialization}. Because of the cyclic
nature of the data series considered, we found it natural to compute
$\widehat{\gamma}_{qn}$ based on estimates of the progression through
these cycles, i.e., by quantizing simple estimates of the instantaneous
phase. To be as fair as possible, the same initial occupancy values
were used across all hidden-state models in a given experiment.
\begin{table}[tp]
\caption{\label{paperd:tab:models}Comparison of models used in the experiments.
Hidden-state models used $M\protect\leq15$ states, with $M=1$ equivalent
to no hidden state. Explicit Markovian dependences were of order $p\protect\leq10$,
with $p=0$ equivalent to no explicit Markovian dependence between
observations.}
\vspace{-1ex}

\noindent \begin{centering}
\tabcolsep=0.15cm\begin{footnotesize}%
\begin{tabular}{|c|c|c|c|}
\hline 
\textbf{Model} & \multicolumn{2}{c|}{\textbf{Time dependence}} & \textbf{Added}\tabularnewline
\cline{2-3} 
\textbf{type} & \textbf{Hidden state} & \textbf{Markov} & \textbf{noise}\tabularnewline
\hline 
\hline 
AR & no & linear AR & Gaussian\tabularnewline
\hline 
HMM & yes & no & Gaussian\tabularnewline
\hline 
AR-HMM & yes & linear AR & Gaussian\tabularnewline
\hline 
KDE-MM & no & \multicolumn{2}{c|}{KCDE, 1 bandwidth}\tabularnewline
\hline 
KDE-HMM & yes & \multicolumn{2}{c|}{KCDE, $M\left(p+1\right)$ bandwidths}\tabularnewline
\hline 
\end{tabular}\end{footnotesize}
\par\end{centering}
\vspace{-2ex}
\end{table}
\begin{figure*}[tp]
\hspace*{\fill}\subfloat[\label{paperd:fig:syntheticgauss}Gaussian-noise driven AR-HMM data.]{\begin{centering}
\includegraphics[width=0.8\columnwidth]{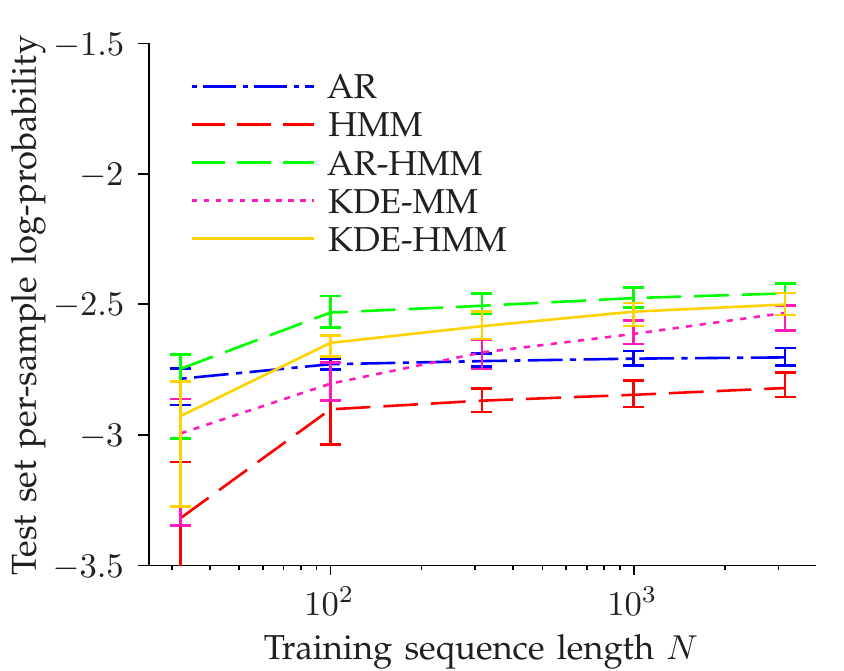}
\par\end{centering}
}\hspace*{\fill}\hspace*{\fill}\subfloat[\label{paperd:fig:syntheticgmm}GMM-driven AR-HMM data.]{\begin{centering}
\includegraphics[width=0.8\columnwidth]{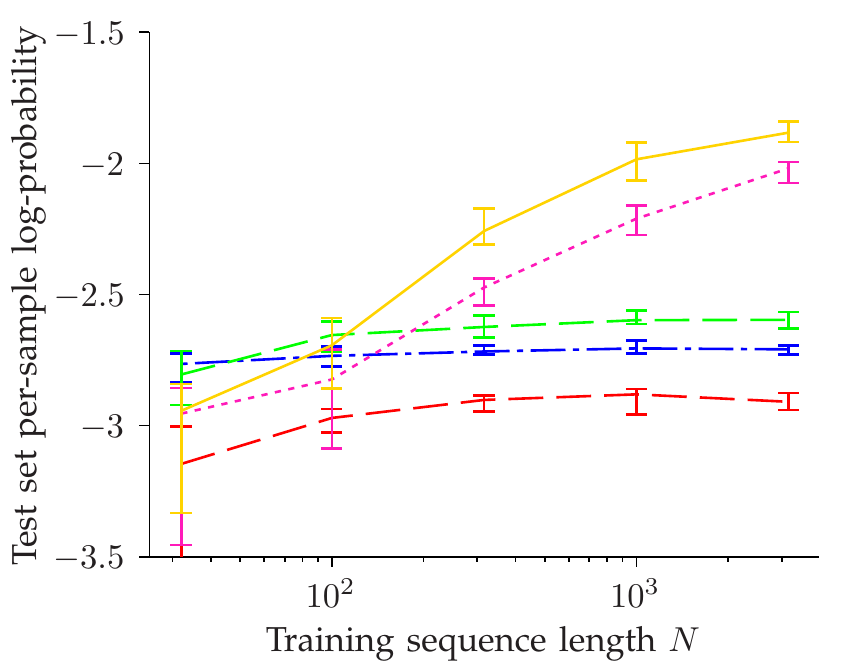}
\par\end{centering}
}\hspace*{\fill}

\caption{\label{paperd:fig:synthetictest}Median held-out set performance of
parametric and nonparametric models from different paradigms as a
function of training dataset size, with error bars at 25\% and 75\%
quantiles. The data was generated by a simple two-state, first-order
AR-HMM, but the two figures differ in the nature of the underlying
driving noise.}
\vspace{-2ex}
\end{figure*}

For the heteroscedastic data considered in this first experiment,
the state (discrete phase) of the variance cycle was estimated based
on the magnitude of process value changes. We then assigned datapoints
to low or high-volatility states using the simple thresholding scheme
\begin{align}
\widehat{\gamma}_{1n}= & \begin{cases}
1 & \textrm{if }n>1\textrm{ and }\left|y_{n}-y_{n-1}\right|\leq\Delta_{\mathrm{th}}\\
\nicefrac{1}{2} & \textrm{if }n=1\textrm{ (since \ensuremath{y_{0}}\ is undefinded)}\\
0 & \textrm{otherwise}
\end{cases}
\end{align}
 and $\widehat{\gamma}_{2n}=1-\widehat{\gamma}_{1n}$, where 
\begin{align}
\Delta_{\mathrm{th}} & =\mathrm{median}\,\left\{ \left|y_{n}-y_{n-1}\right|\right\} _{n=2}^{N}\textrm{.}
\end{align}
 Similar to KDE-HMMs, the single KDE-MM bandwidth parameter was initialized
based on the normal reference rule \cite{scott1992a,bowman1997a}.

For each dataset generated, each of the five models we consider was
initialized and subjected to 500 iterations on the training data (excluding
linear AR models which do not require iteration). The trained models
were then asked to compute the log-probability of the corresponding
set of separate but similarly generated validation data. Figs.\ \ref{paperd:fig:syntheticgauss}
and \ref{paperd:fig:syntheticgmm} illustrate the median validation
set performance for different training data sizes $N$ on the two
data sources, with one curve for each method, surrounded by 25\% and
75\% quantile intervals. Higher log-probabilities are better; an upper
bound on expected performance is given by the differential entropy
rate of the data source.

\subsection{Analysis}

Despite the simplistic choice of weights $w_{qn}$, KDE-HMMs were
capable of learning good models of autoregressive hidden-state processes
driven both by Gaussian noise as well as by more general GMM noise,
given a sufficient amount of data. On the other hand, neither pure
Markovian nor pure hidden-state models were a good fit for GMM-driven
data, though the models are likely to improve in asymptotic performance
if more states or higher orders were to be considered.

Unsurprisingly, parametric models converged faster than KDE-HMMs,
but provided inferior asymptotic performance unless the data was generated
by a model within their particular parametric class. This is why Gaussian
AR-HMMs show good performance in the first figure, where $U_{t}$
happens to be Gaussian, but not in the second, where $U_{t}$ follows
a GMM. In fact, all non-KDE models show reduced asymptotic performance
for the second dataset, even though the underlying process has lower
entropy rate, making it, in principle, easier to predict. The issue
is that the parametric models place the majority of their probability
mass in regions near the next-step conditional mean $\mathbb{E}\left(X_{t+1}\mid\underline{x}_{1}^{t}\right)$,
where the true pdf is quite small.

\section{Experiments on Real-Life Data\label{paperd:sec:realexperiments}}

In our second set of experiments, we investigated the capabilities
of KDE-MMs and KDE-HMMs for describing several challenging natural
processes of interest in nonlinear prediction, and compared against
baseline models. The results affirmed the advantages of KDE-based
techniques. Moreover, the lag-dependent bandwidths and hidden-state
memory of KDE-HMMs were shown to improve on KDE-MM performance for
all datasets.

\subsection{Data Series}

For our first real-world application, we considered the laser data
(dataset A and its continuation) from the Santa Fe time-series prediction
contest described in \cite{weigend1994a}. This data was also used
in \cite{ralaivola2005a} and \cite{kallas2011a}, for instance. The
data consists of integer-quantized intensity measurements from a laser
in a chaotic state. To avoid issues with degenerate likelihoods due
to coinciding points (a problem with all latent-variable models considered)
uniform noise over $\left(-\nicefrac{1}{2},\,\nicefrac{1}{2}\right)$
was added to the data.

\begin{figure*}[tp]
\hspace*{\fill}\subfloat[\label{paperd:fig:laseroverview}Excerpt from laser data.]{\begin{centering}
\includegraphics[width=0.8\columnwidth]{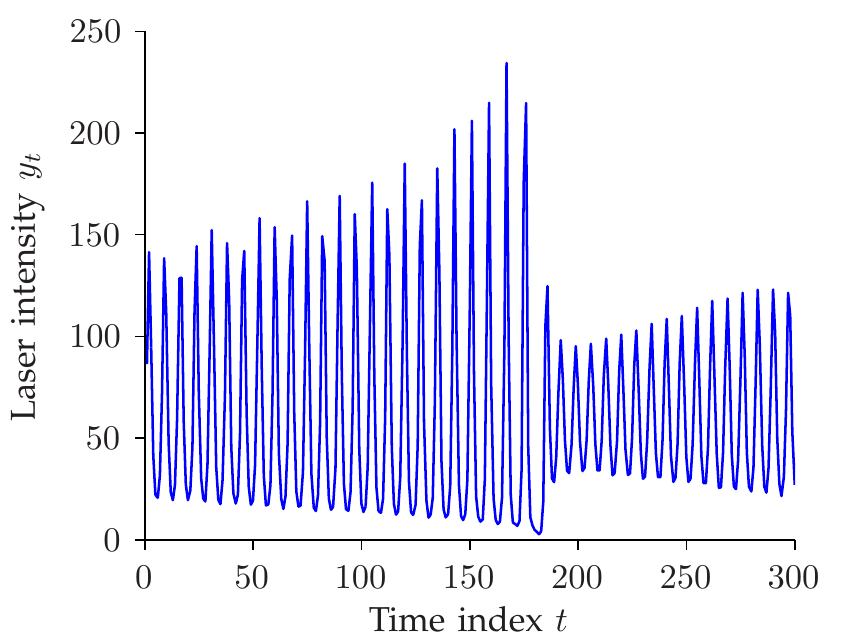}
\par\end{centering}
}\hspace*{\fill}\hspace*{\fill}\subfloat[\label{paperd:fig:ecgoverview}Excerpt from ECG data.]{\begin{centering}
\includegraphics[width=0.8\columnwidth]{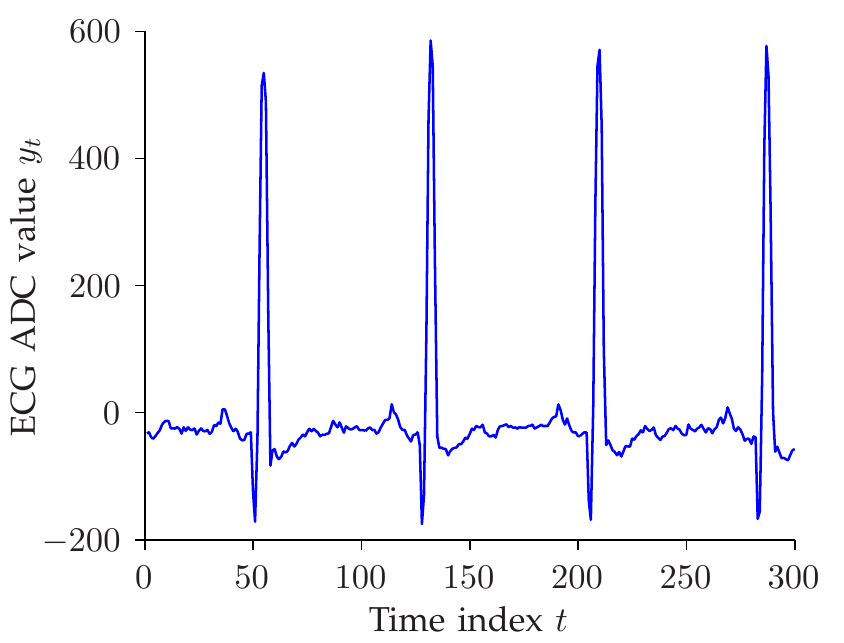}
\par\end{centering}
}\hspace*{\fill}\\
\hspace*{\fill}\subfloat[\label{paperd:fig:lasermarkovlogprob}Markov-model performance on
laser data.]{\begin{centering}
\includegraphics[width=0.8\columnwidth]{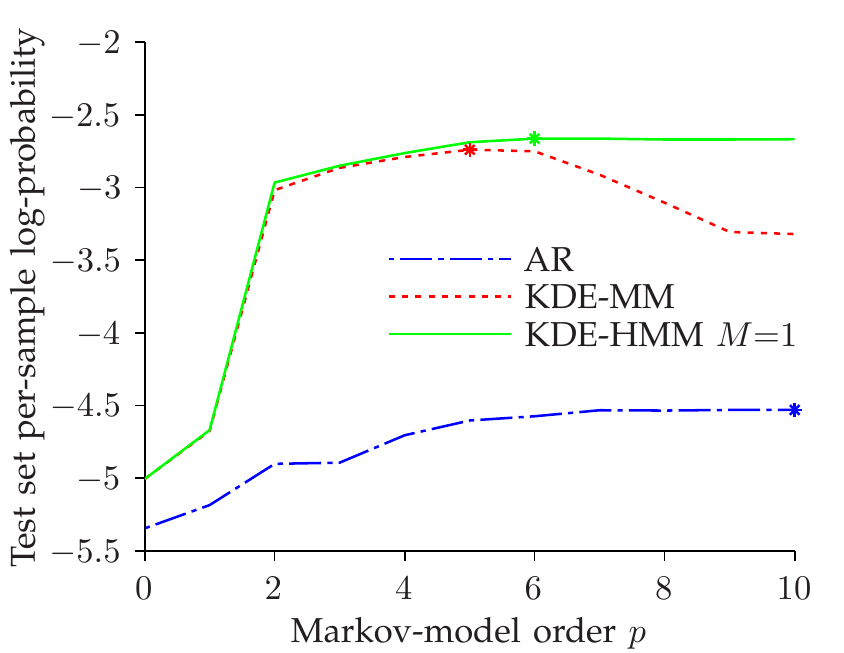}
\par\end{centering}
}\hspace*{\fill}\hspace*{\fill}\subfloat[\label{paperd:fig:ecgmarkovlogprob}Markov-model performance on ECG
data.]{\begin{centering}
\includegraphics[width=0.8\columnwidth]{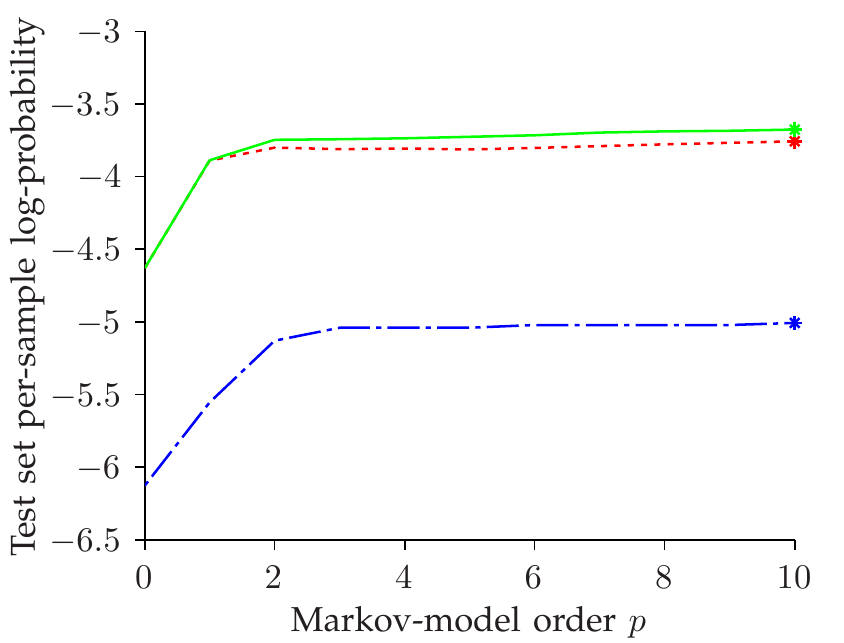}
\par\end{centering}
}\hspace*{\fill}\\
\hspace*{\fill}\subfloat[\label{paperd:fig:laserhmmlogprob}Hidden Markov model performance
on laser data.]{\begin{centering}
\includegraphics[width=0.8\columnwidth]{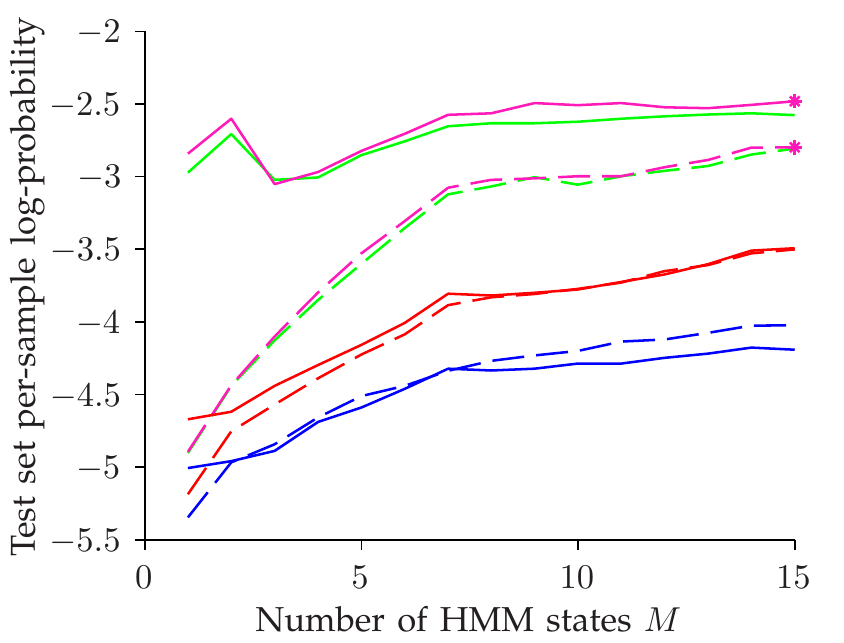}
\par\end{centering}
}\hspace*{\fill}\hspace*{\fill}\subfloat[\label{paperd:fig:ecghmmlogprob}Hidden Markov model performance on
ECG data.]{\begin{centering}
\includegraphics[width=0.8\columnwidth]{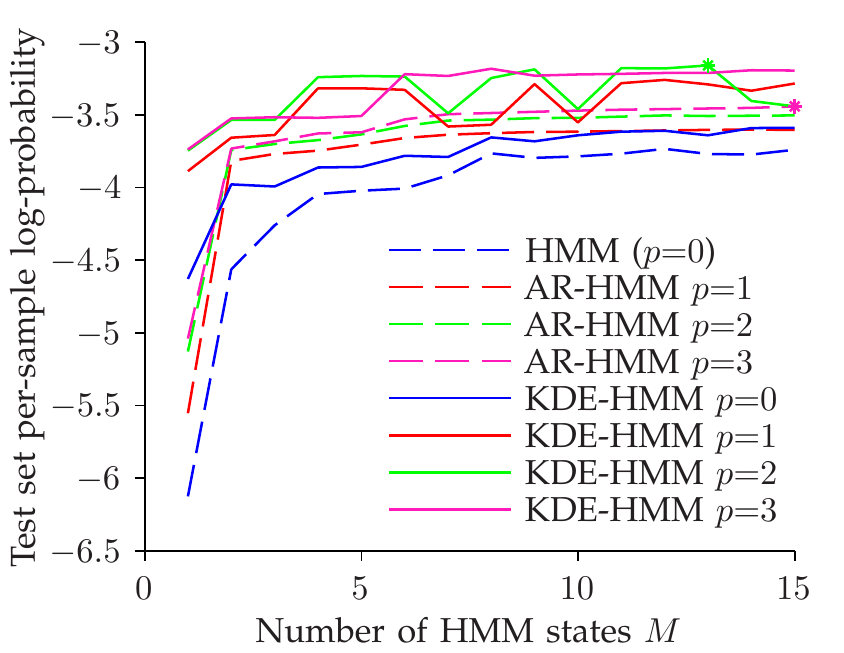}
\par\end{centering}
}\hspace*{\fill}

\caption{\label{paperd:fig:logprobtest}Training series and held-out set (natural)
log-probability of various models for real-life data experiments.
Standard AR models are dash-dotted, HMMs and AR-HMMs are dashed, KDE-MMs
are dotted, and KDE-HMMs are drawn solid. Asterisks mark the best
model of each kind. In the last two figures, colours reflect model
order $p$.}
\end{figure*}

A plot of the first 300 points of the laser data is provided in Fig.\ \ref{paperd:fig:laseroverview}.
The series shows oscillations, about seven samples long, that slowly
increase in magnitude, only to eventually fizzle out and start over
again.

The second data series consisted of raw AD-converter values from an
ECG signal sampled at 128 Hz, specifically series 16265 in the MIT-BIH
Normal Sinus Rhythm Database from PhysioNet \cite{goldberger2000a}.
This data was previously considered in \cite{kallas2013a}. Like the
laser data, uniform noise was added to counteract quantization effects.
A plot of the first 300 points of the data is provided in Fig.\ \ref{paperd:fig:ecgoverview}.
The data has the characteristic ECG shape, showing regular, distinct
pulses with lesser activity in between.

For both datasets, the first $N=3000$ points were used for training
and the $N$ subsequent points for validation.

\subsection{Markov-Model Experiment}

A number of experiments were performed on each dataset. In a first
application, three different types of Markov models were compared.
Specifically, linear AR-models, KDE-MMs, and KDE-HMMs with a single
state (so no long-range memory) of orders $p=0$ (corresponding to
a conditional independence assumption) through $10$ were fitted to
the training data. The models were initialized as in Section \ref{paperd:sub:initialization},
but with $\gamma_{1t}=1$ since only a single state was used. Due
to the low dimensionality of the parameter space, Matlab \texttt{fminunc}
sufficed for optimizing KDE bandwidths.

After training, the models were asked to assess the probability of
the held-out validation data, with results as shown in Figs.\ \ref{paperd:fig:lasermarkovlogprob}
and \ref{paperd:fig:ecgmarkovlogprob}. Higher probabilities are preferred,
as before. Apart from stochastic effects, increased held-out set log-probability
directly implies a proportional reduction in KL-divergence, cf.\ (\ref{paperd:eq:kldiv}).

The results of the Markov-model experiment highlight the power of
nonparametric approaches for nonlinear datasets. KDE-MMs greatly outperform
linear AR models of the same order. Single-state ($M=1$) KDE-HMMs,
which additionally offer lag-dependent bandwidths, are better still,
showing minor improvements over KDE-MMs everywhere except at $p=0$,
where the two models coincide. Since the nonparametric models use
the information provided by the context more efficiently, their performance
saturates faster, and at a higher level, as $p$ increases.

The fact that general nonparametric density estimation rapidly becomes
more data-demanding in higher dimensions \cite{silverman1986a} could
be cause for concern, as our models conceptually are based on estimating
$f_{\underline{X}_{1}^{p+1}}$ using KDE. However, the experiments
show that high dimensionalities are not as problematic as may have
been anticipated, and KDE-HMMs consistently work well all the way
up to $p+1=11$-dimensional contexts. One explanation is that KDE-HMMs
are capable of automatic relevance determination: by setting $h_{ql}$
large, the corresponding context values $y_{n-l}$ exert negligible
influence on the distribution of $y_{n}$. Models can thus ignore
non-informative lags. (Linear AR-models have similar abilities to
exclude variables from consideration.) The learned laser data bandwidth
parameters were found to be very wide for $l>6$, consistent with
this hypothesis. KDE-MMs, in contrast, are constrained to use the
same bandwidths for all lags and may suffer when the context includes
variables of little predictive value, explaining the performance roll-off
after $p=6$ in Fig.\ \ref{paperd:fig:lasermarkovlogprob}.

\subsection{Hidden Markov Model Experiment}

Next, we investigated the benefits of adding a hidden state to the
various models. To do this, we trained both KDE-HMMs and regular HMMs
with Gaussian autoregressive outputs (AR-HMMs \cite{shannon2009a})
of orders $p=0$ through $3$, with $M$ states, $M\in\left\{ 1,\,\ldots,\,15\right\} $,
for each $p$. When $p=0$, the AR-HMMs reduce to ordinary HMMs with
Gaussian output distributions, while the KDE-HMMs resemble the KDE/HMMs
in \cite{piccardi2007a}. Many Markov models in the previous section
are special cases for $M=1$.

All the participating models were initialized using the same scheme
as in the previous experiments. Like in Section \ref{paperd:sub:synthsetup},
initial state occupancies $\widehat{\gamma}_{qt}$ were based on the
estimated cycle state. To estimate the continuous instantaneous phase
for initialization, the peak of each oscillation cycle in the data
was extracted; by assuming consecutive peaks were separated by a phase
difference of $2\pi$, instantaneous phase values $\widehat{\varphi}_{t}$
for all $t$ could be interpolated using cubic splines. The initial
occupancies were computed as a soft quantization of this phase, 
\begin{align}
\widehat{\gamma}_{qt} & =\max\left(0,\,1-\frac{M}{2\pi}\min\left(\Delta\widehat{\varphi}_{qt},\,2\pi-\Delta\widehat{\varphi}_{qt}\right)\right)\\
\Delta\widehat{\varphi}_{qt} & =\left(\widehat{\varphi}_{t}-2\pi\frac{q-1}{M}\right)\mathrm{mod}\,2\pi\textrm{.}
\end{align}
 This satisfies $\sum_{q}\widehat{\gamma}_{qt}=1$ for all $t$, as
required, and provides a sparse initialization where most $\widehat{\gamma}_{qt}$-values
are zero. The setup promotes a hidden-state memory that tracks the
current position in the cycle, for instance biasing ECG pulses to
appear at regular intervals.

Following training, the log-probability of the held-out dataset was
computed under each model, with results as in Figs.\ \ref{paperd:fig:laserhmmlogprob}
and \ref{paperd:fig:ecghmmlogprob}. A number of trends are evident
in these figures. First, combining the local continuity of Markov
models with the hidden state variable of HMMs as suggested in \ref{paperd:sub:combinationmodels}
increased the accuracy of both the parametric and the nonparametric
techniques. For a fixed $p$, increasing $M$ typically improved performance
across all models. This is especially clear when going from $M=1$
to $M=2$, but we also note that hidden-state model performance at
high $M$ exceeds the asymptotic levels attained by the Markov-models
in Figs.\ \ref{paperd:fig:lasermarkovlogprob} and \ref{paperd:fig:ecgmarkovlogprob},
even though these used considerably higher $p$-values. Conversely,
given a fixed $M$, increasing $p$ often proved beneficial, particularly
at low orders. In other words, more parameters tended to help, both
for the linear, parametric baseline models and for KDE-HMMs.

Most importantly, even though the weaker linear AR models obviously
had the most to gain from adding a hidden state, the figures show
that KDE-HMMs provided greater accuracy than parametric models of
similar order $p$ and state-set size $M$. This is true for all graphs
in Fig.\ \ref{paperd:fig:ecghmmlogprob}, and for the laser results
in \ref{paperd:fig:laserhmmlogprob} once $p\geq2$.\footnote{Given how regular the laser data is, having $p\geq2$ context values
is highly informative, as it not only reveals the most recent value
of the process, but also in which direction it is changing and how
rapidly.}

For the ECG data in \ref{paperd:fig:ecghmmlogprob}, KDE-HMM behaviour
at $p\geq1$ appears somewhat inconsistent, in that performance seems
to fluctuate between two distinct levels, one noticeably better than
the other. This suggests that the maximization procedure, being sensitive
to initialization, sometimes converged to local optima of inferior
quality. Despite the fluctuations, KDE-HMMs always outperform their
matching AR-HMM counterparts for all $\left(M,\,p\right)$ combinations
in the figure.

As discussed in Section \ref{paperd:sub:hmms}, the addition of a
hidden state helps both by partitioning the context-space and by providing
long-range memory. The results show that such partitioning is useful
for all model types, but especially benefits weaker models like Gaussian,
linear AR, by enabling a crude form of nonlinearity where different
parts of the data are described by different models. KDE-HMMs further
improve on parametric hidden-state approaches by relaxing assumptions
on distributions and within-state linearity.

\begin{figure}[tp]
\begin{centering}
\includegraphics[width=0.8\columnwidth]{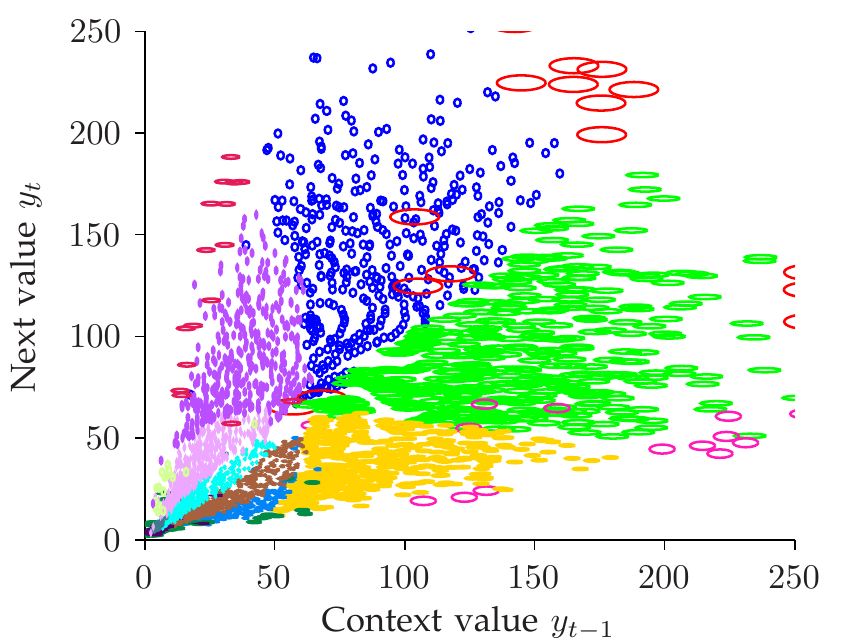}
\par\end{centering}
\caption{\label{paperd:fig:kdehfromq}Scatter plot illustrating state-dependent
kernel widths $\widehat{h}_{\widehat{q}_{n}l}$ of KDE-HMM trained
on the laser data. Colours are based on datapoint state assignment
$\widehat{q}_{n}=\mathrm{argmax}_{q}\gamma_{qn}$. The swirly pattern
of the data is due to structure in the strange attractor.}
\vspace{-2ex}
\end{figure}

Data partitioning also gives KDE-HMMs an edge over KDE-MMs, since
it allows for context-dependent bandwidths. This is illustrated by
Fig.\ \ref{paperd:fig:kdehfromq}, displaying a scatter plot of the
laser training data, with each datapoint coloured by state assignment
($\widehat{q}_{n}=\mathrm{argmax}_{q}\gamma_{qn}$) for the best-performing
KDE-HMM ($M=15$, $p=3$). Drawing each point as an ellipse centred
on $\left(y_{n-1},\,y_{n}\right)$ with axes dimensioned according
to $\widehat{h}_{\widehat{q}_{n}l}$, we can confirm that the trained
model adaptively uses wider bandwidths in regions where data is sparse,
but employs narrower bandwidths to bring out detail in more concentrated
regions. A KDE-MM, in contrast, cannot do this, and is forced to use
a wide compromise bandwidth throughout.

As anticipated, parametric models were faster to work with throughout
the experiments, since their computational demands per training iteration
are linear in $N$, whereas KDEs with pseudo-likelihood maximization
scale as $\mathcal{O}\left(N^{2}\right)$. We envision that additional
approximations, particularly in the style of \cite{holmes2007a},
will be essential to fast computation in large-scale applications.
We also expect KDE-HMM accuracy would improve further, and the top-left
dip in the performance curves in Fig.\ \ref{paperd:fig:laserhmmlogprob}
could be eliminated, if the weights $w_{qn}$ could be trained efficiently.

\section{Conclusions and Future Work\label{paperd:sec:conclusion}}

We have described KDE-MMs and KDE-HMMs as nonlinear, nonparametric
models of stochastic time series. Unlike traditional parametric approaches,
these models can represent a broad class of continuous-valued stochastic
processes, including all Markov processes. We also detailed how the
models can be trained, and demonstrated good modelling performance
in applications to synthetic and natural data.

For future applications it is especially compelling to investigate
the use of KDE-HMMs in speech synthesis. The KDE-HMM data-generation
mechanism, which in essence concatenates datapoints from the training
material in a context-sensitive manner, is evocative of concatenative
speech synthesis as popularized by \cite{hunt1996a}, except that
KDE-HMMs also are fully probabilistic. This is promising, seeing that
synthesis techniques based on probabilistically-guided exemplar concatenation
have produced leading results in recent speech-synthesis challenges
\cite{qian2010hmm,ling2012a}.

\bibliographystyle{IEEEtran}
\bibliography{kde-hmm_2015_refs}

\end{document}